\documentclass[10pt,twocolumn,letterpaper]{article}
\usepackage[dvipsnames]{xcolor}

\usepackage{wacv}      
\usepackage{times}
\usepackage{epsfig}
\usepackage{graphicx}
\usepackage{amsmath}
\usepackage{amssymb}
\usepackage{booktabs}
\usepackage{url}
\usepackage[accsupp]{axessibility}  

\definecolor{darkred}{rgb}{0.7,0.1,0.1}
\definecolor{darkgreen}{rgb}{0.1,0.7,0.1}
\definecolor{cyan}{rgb}{0.7,0.0,0.7}
\definecolor{dblue}{rgb}{0.2,0.2,0.8}
\definecolor{maroon}{rgb}{0.76,.13,.28}
\definecolor{burntorange}{rgb}{0.81,.33,0}

\newcommand*{\ShowNotes}{}

\ifdefined\ShowNotes
  \newcommand{\colornote}[3]{{\color{#1}\bf{#2: #3}\normalfont}}
\else
  \newcommand{\colornote}[3]{}
\fi

\def\wacvPaperID{1310} 
\def\confName{WACV}
\def\confYear{2024}

\usepackage[pagebackref,breaklinks,colorlinks]{hyperref}

\usepackage[capitalize]{cleveref}
\crefname{section}{Sec.}{Secs.}
\Crefname{section}{Section}{Sections}
\Crefname{table}{Table}{Tables}
\crefname{table}{Tab.}{Tabs.}
\definecolor{shape_green}{rgb}{0.1,0.7,0.1}
\definecolor{period_purple}{rgb}{0.7,0.0,0.7}

\begin{document}

\teaser{
\centering
\begin{tabular}{ c c c c}
    \includegraphics[width=0.12\textwidth]{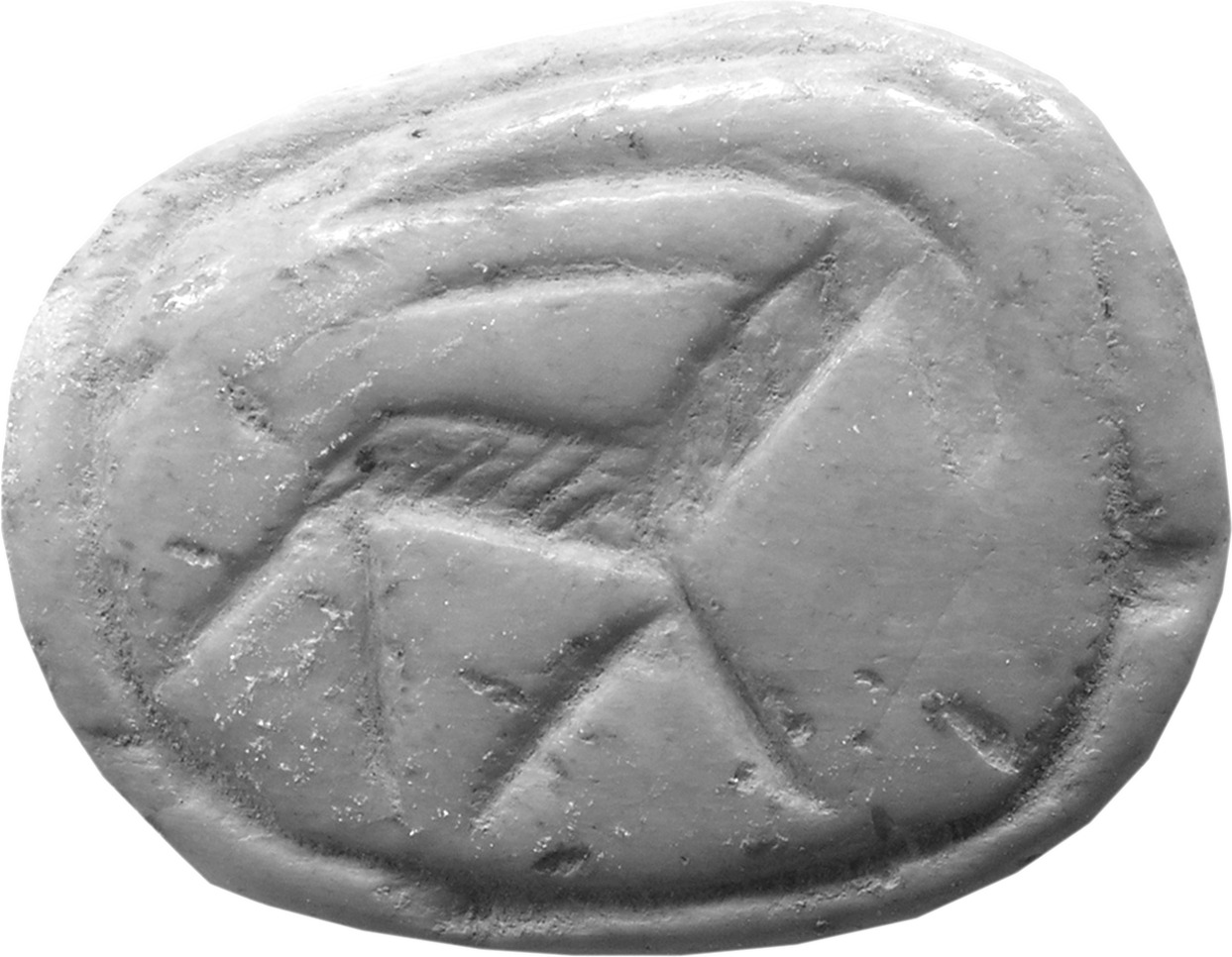} &
 \includegraphics[width=0.25\textwidth]{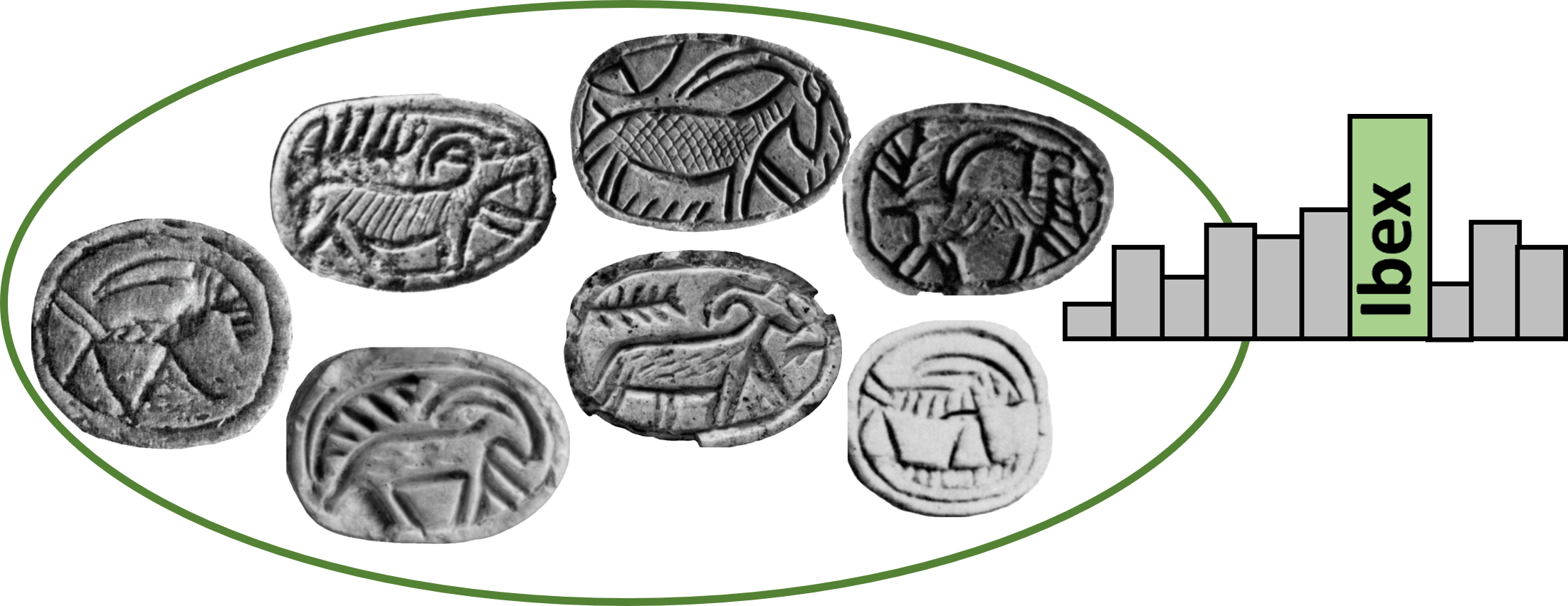} & 
 \includegraphics[width=0.22\textwidth]{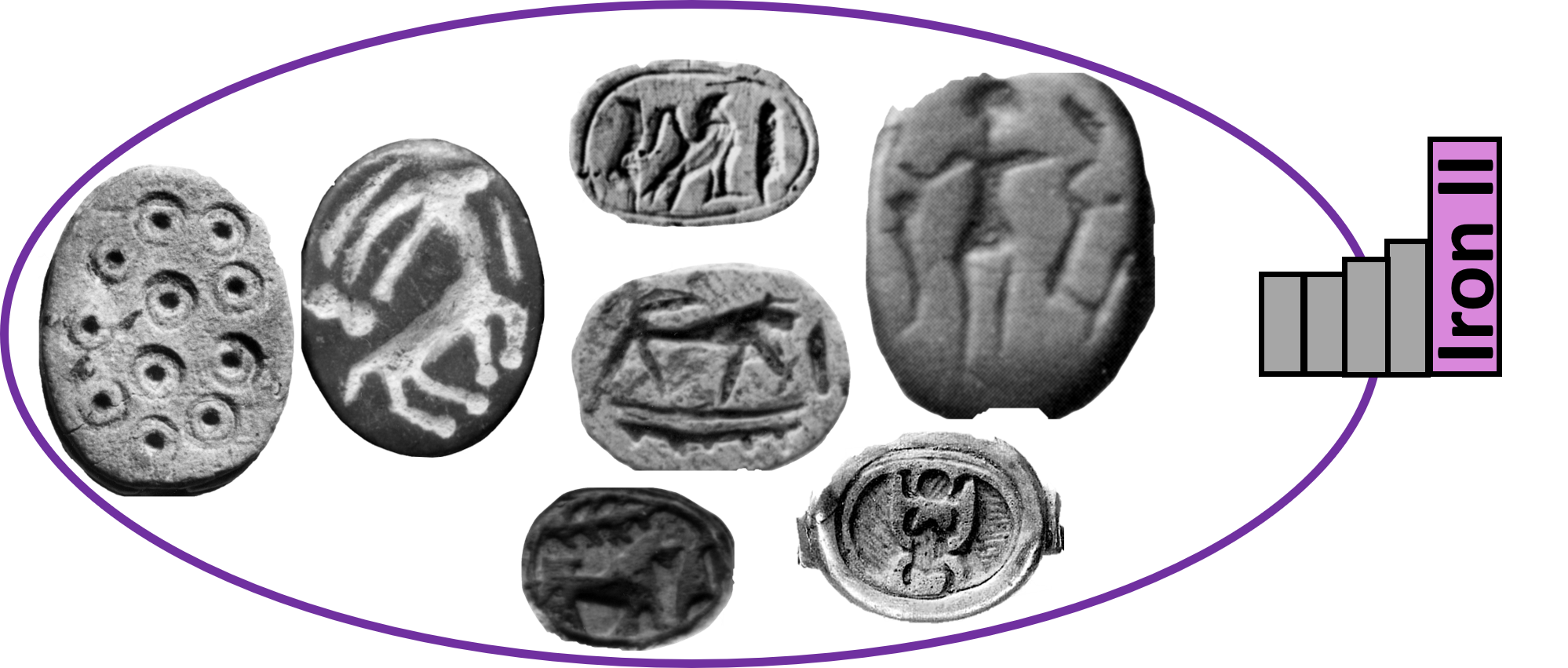} &
  \includegraphics[width=0.12\textwidth]{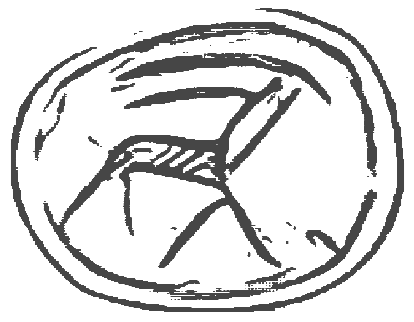} \\
 \includegraphics[width=0.12\textwidth]{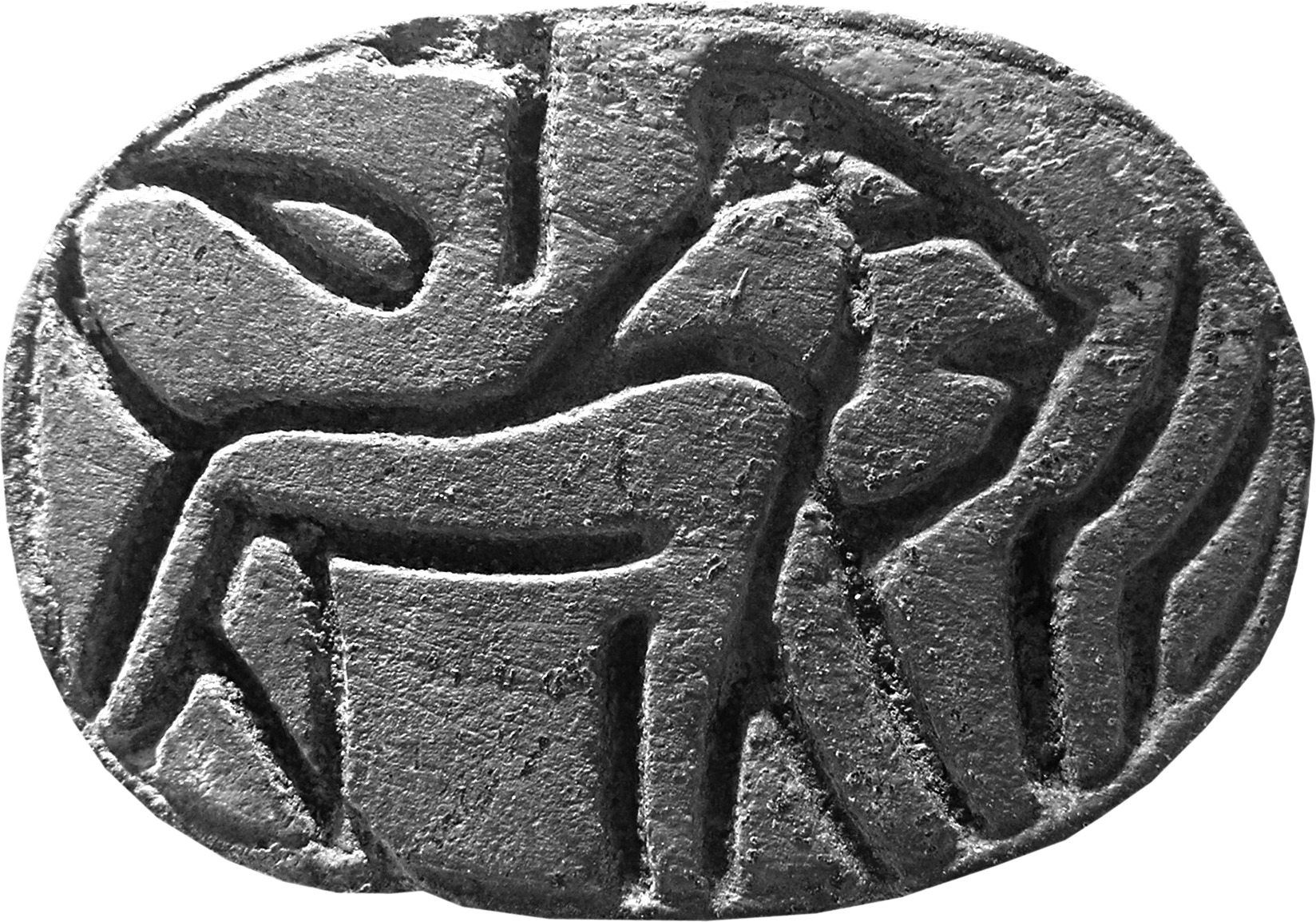} &
 \includegraphics[width=0.25\textwidth]{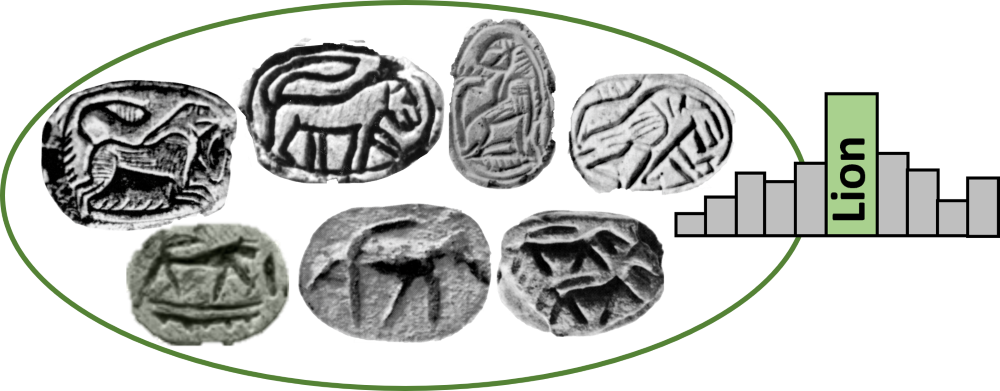} &
 \includegraphics[width=0.22\textwidth]{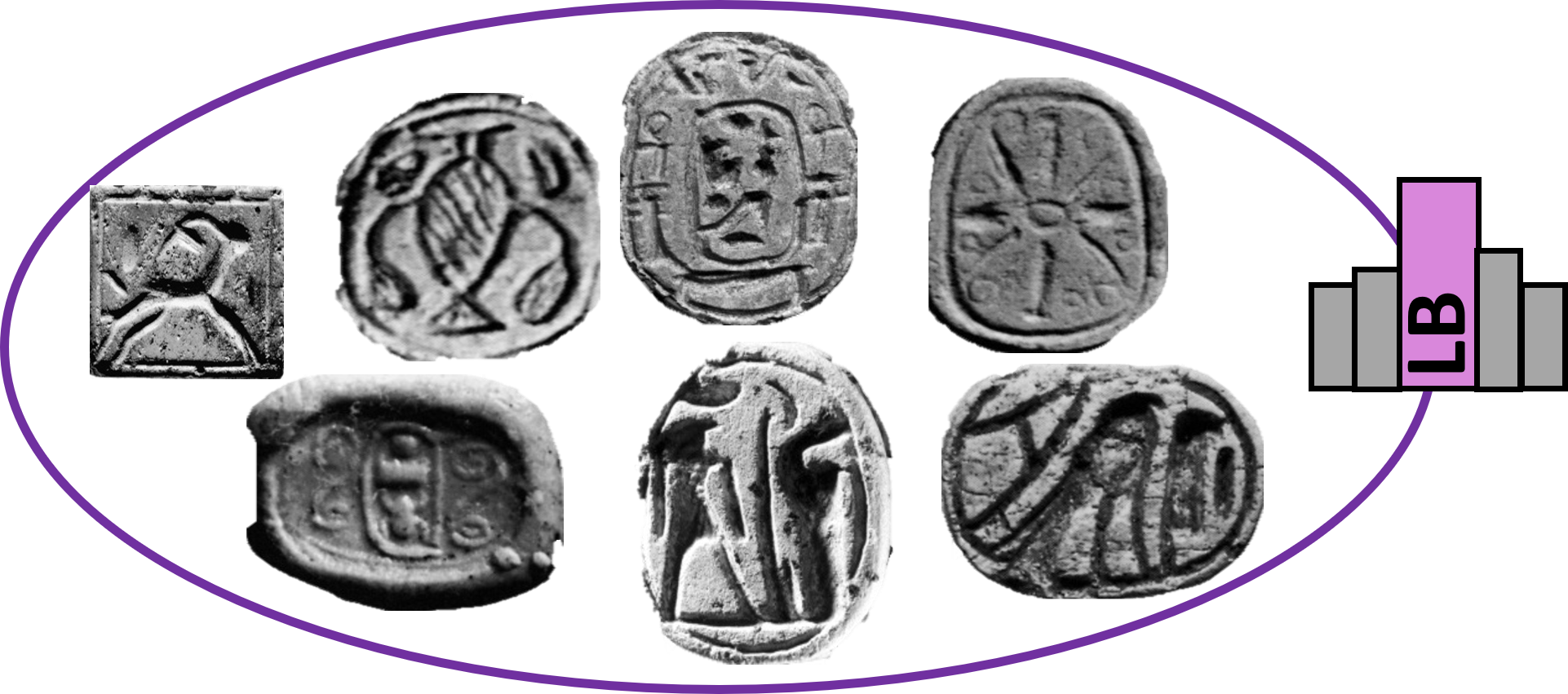} & 
 \includegraphics[width=0.12\textwidth]{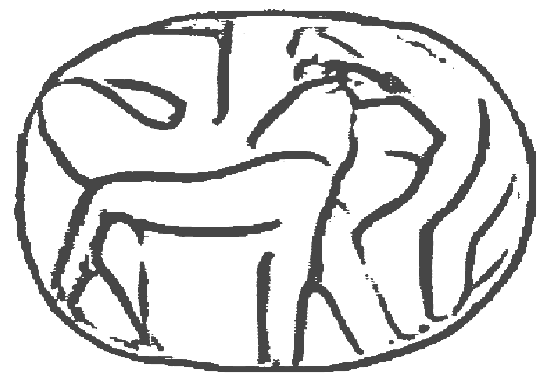} \\
(a) Input image & (b) \textcolor{shape_green}{Classification by shape} & (c) \textcolor{period_purple}{Classification by period} & (d) Generation \\
\end{tabular}
\caption{{\bf Analyzing archaeological artifacts.}
Images of archaeological artifacts are extremely challenging to analyze, since they are eroded, broken and stained~(a). 
Our model manages to classify these artifacts not only by \textcolor{shape_green}{shape}~(b), but also by \textcolor{period_purple}{historical period}~(c).
In this figure, each ellipse bounds artifacts from the same class, demonstrating the great diversity within a class. 
Furthermore, our model generates a drawing of the given artifact~(d), which is a standard way of archaeological documentation.
The top row shows an \textcolor{shape_green}{{\em Ibex}} dated to the \textcolor{period_purple}{{\em Late Iron}}, whereas the bottom row shows a \textcolor{shape_green}{{\em Lion}} dated to the \textcolor{period_purple}{{\em Late Bronze}}.
}
\label{fig:teaser}
}

\title{ArcAid: Analysis of Archaeological Artifacts using Drawings}
\author{Offry Hayon\\
Technion\\
 {\tt\scriptsize offryh@gmail.com}
\and
Stefan Münger\\
U. of Bern\\
 {\tt\scriptsize stefan.muenger@unibe.ch}
\and
Ilan Shimshoni\\
U. of Haifa\\
{\tt\scriptsize ishimshoni@is.haifa.ac.il}
\and
Ayellet Tal\\
Technion\\
{\tt\scriptsize ayellet@ee.technion.ac.il}
}

\maketitle
\thispagestyle{empty}

\begin{abstract}
Archaeology is an intriguing domain for computer vision.
It suffers not only from shortage in (labeled) data, but also from highly-challenging data, which is often extremely abraded and damaged. 
This paper proposes a novel semi-supervised model for classification and retrieval of images of archaeological artifacts.
This model utilizes unique data that exists in the domain---manual drawings made by special artists.
These are used during training to implicitly transfer the domain knowledge from the drawings to their corresponding images, improving their classification results.
We show that while learning how to classify, our model also learns how to generate drawings of the artifacts, an important documentation task, which is currently performed manually.
Last but not least, we collected a new dataset of stamp-seals of the Southern Levant.
Our code\footnote{\url{https://github.com/offry/Arc-Aid}} and dataset\footnote{\url{https://cgm.technion.ac.il/arcaid/}} are publicly available.
\end{abstract}

\section{Introduction}
\label{sec:intro}

Archaeology benefits society through understanding of the past and is acknowledged worldwide as a major research field.
Advances in computer vision  may be harnessed to the task in order to automatize some aspects of the study of archaeological findings (artifacts).
For instance, a core task in the domain is to look for similar artifacts, which may reveal relations, commerce and connections between countries and cultures.
The current practice is to leaf through thousands of pages in site reports.
Instead, performing this task (and others) utilizing vision methods could take minutes.
These methods are essential not only because the number of artifacts is large and the number of experts is small, but also because datasets are distributed all over the world. 

%
This, however, is an intriguing task, as  the archaeological domain exposes the limits of current computer vision techniques, due to  several unique properties.
First, there is a shortage of labeled data, since  labelling must be performed by archaeological experts.
Second, many archaeological artifacts are preserved in poor state of condition, eroded or broken, which differs from that of standard natural images.
Third, since the artifacts are hand crafted, the consistency between different items of the same class is relatively weak.

A major task in archaeology is to classify artifacts by different criteria.
The few classification methods in the domain exhibit good results, however they mostly focus on classes that have small variety within each class (for instance, similar coins with varied state of preservation)~\cite{resler2021deep, cooper2020learning, ma2020classification, chetouani2020classification, anwar2021deep, canul2018classification, barucci2021deep}.
Our goal is broader: classifying a given artifact, where images in each class may vary greatly.
This is either because they were produced in different periods or by different artists (e.g. considerably different  lions and ibexes shown in Figure~\ref{fig:teaser}(b)) or because they are clustered by periods, even though the appearance of the shapes from the same period inherently differ (Figure~\ref{fig:teaser}(c)).

Since we do not have access to the real artifacts, which are often kept in store rooms of archaeological services, we focus on the visual documentation of these artifacts.
One obvious such documentation is images, which is our input.
Oftentimes, the images of the artifacts are accompanied by illustrations, made by trained  drafts persons. 
Though the drawings and the photos are not necessarily aligned and the drawings are not exact depictions of the images, some features of the artifacts look clearer and more enhanced in the drawings.
Thus, we propose to use them during training.

We introduce a novel semi-supervised approach for classifying archaeological artifacts. 
During training we utilize unlabeled pairs of drawings and images, together with a smaller number of labeled pairs.
At inference, however, \underline{only an image} is given as input.
Our approach addresses the unique domain's challenges---a small dataset, the poor state, the lack of consistency between artifacts in the same class, and similar objects in  different classes (which require expertise).
Specifically, as demonstrated in Figure~\ref{fig:teaser}(b)-(c), our model classifies both by shape and by period.
In these examples, the given image (at inference) is not ideal in terms of quality.
Yet, our model classifies the images correctly according to both classification types.

\begin{figure}[t]
\begin{center}
\begin{tabular}{c c c c}
 \includegraphics[height=1.5cm]{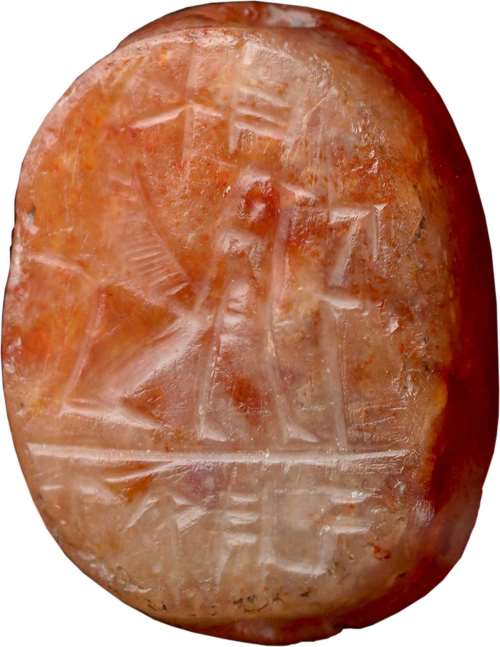} & \includegraphics[height=1.3cm]{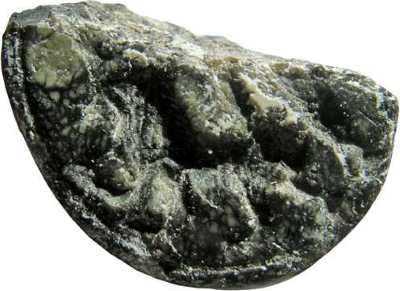} & 
 \includegraphics[height=1.4cm]{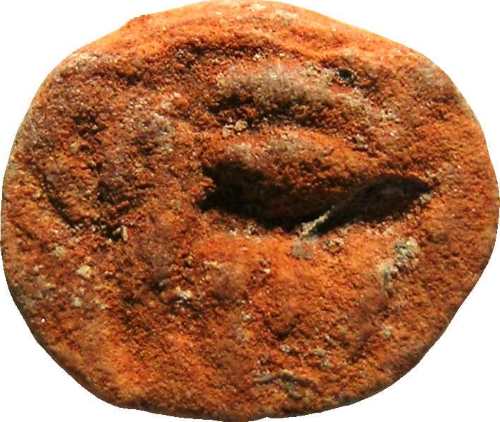} & 
 \includegraphics[height=1.4cm]{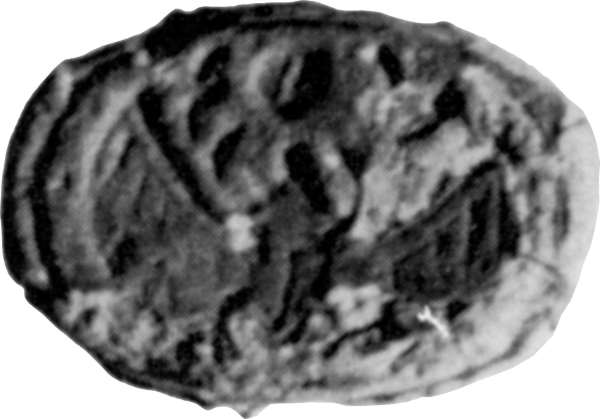} \\  
 \includegraphics[height=1.5cm]{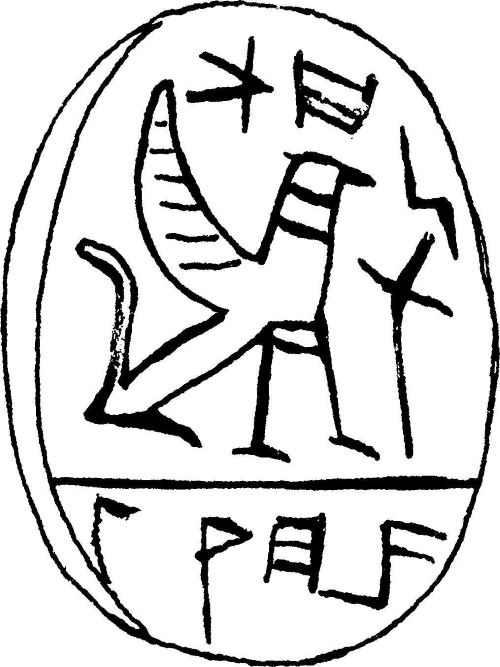} & \includegraphics[height=1.2cm]{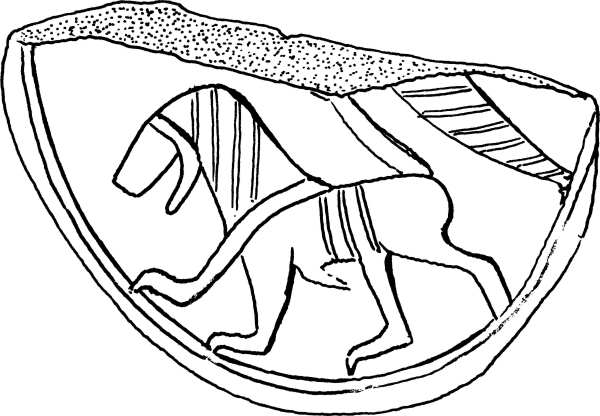} & 
 \includegraphics[height=1.3cm]{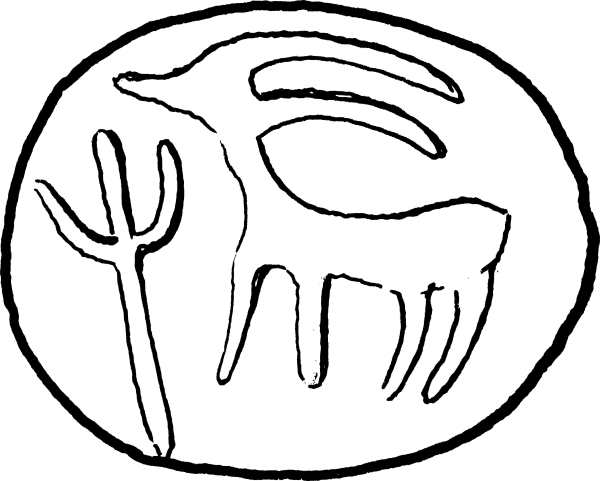} & 
 \includegraphics[height=1.3cm]{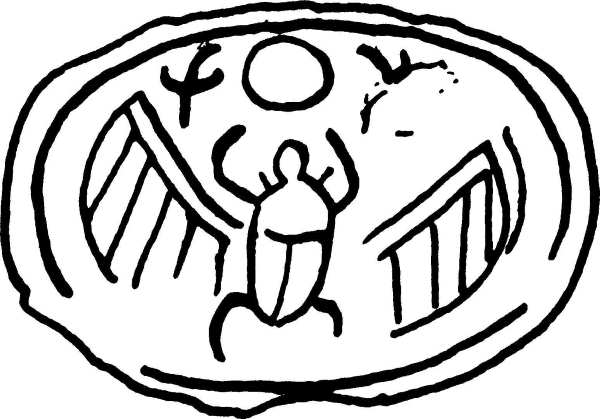} \\  
 (a) Bird & (b) Lion & (c) Ibex & (d) Beetle    
\end{tabular}
\caption{{\bf Input to training.} The edges of the drawings are clear and  complete, comparable to their counterpart in the  images.
Moreover,  the details differ and the pairs are misaligned.}
\label{fig:pairs}
\end{center}
\end{figure}

Our method is based on a key observation that although the drawings are not exact edge detections of the photos,  utilizing them during training is beneficial.
This is so since they both represent the same main features of the artifact, so the global features found in both are similar, but clearer to detect in a drawing due to the state of the artifact  (Figure~\ref{fig:pairs}).
Forcing similarity between the embeddings of images and drawings contributes to the representation learning.

During training we solve an additional  task---drawing generation---with unlabeled image-drawing pairs.
Currently, since drawing generation requires special artists, it is done for very few selected artifacts, rather than to the whole data. 
We present SoTA results in  classification,  retrieval, and image-to-drawing generation in our domain.


Last but not least, we present a novel dataset, {\em Corpus of the Stamp-seals of the Southern Levant (CSSL).}
It contains scarabs and other seals from Egypt and the Southern Levant ($1750$-$330$ BCE), classified by experts both by shape and period.
This is an important contribution, since archaeological datasets are rare in general, and in particular datasets of paired images and drawings.

Hence, this paper makes three contributions. 
\begin{enumerate}
\vspace{-0.04in}
    \item 
    It presents a semi-supervised method for  multi-modal learning of paired drawings \& photos in archaeology.
\vspace{-0.04in}
    \item
     It introduces a model for image classification and retrieval of scarabs, with respect to the shape or the period, jointly with the ability for drawing generation.
\vspace{-0.04in}
    \item
    We collected a new dataset of images and drawings of decorated scarabs, together with classifications and retrieval benchmarks according to both shape and period. 
\end{enumerate}

\section{Related work}
\label{sec:related work}
\noindent
{\bf Computer vision for archaeology.}
Most of the papers that develop vision techniques for the archaeological domain address one of three tasks:
(1)~documentation, where the goal is to either generate 3D curves~\cite{kolomenkin2009edge, lawonn2016visualization} or to extract  reliefs~\cite{choi2021deep, choi2020relief, zeppelzauer2016interactive, 10.1145/3465629};
(2)~restoration, where the attempt is to restore the way the artifact looked before it was damaged, such as in hole completion~\cite{sipiran2018completion, lengauer2022context, sipiran2022data}, or in reassembly~\cite{derech2021solving, cantoni2020javastylosis, 10.1145/3569091, 10.1145/3417711};
(3)~location, where the attempt is to locate the artifact in time and space through classification~\cite{barucci2021deep, resler2021deep, anwar2021deep, ma2020classification, itkin2019computational, 5539841, cooper2020learning, shen2021large} or retrieval~\cite{resler2021deep, sipiran2021shrec, karl2022advances, anwar2020image, 7816648, shen2021large}.

Our focus in this paper is on classification and retrieval of images of archaeological artifacts.
In~\cite{anwar2021deep} a new architecture, {\em CoinNet}, is presented, which performs classification by shape (decoration).
Results are presented on a Roman coins dataset.
In~\cite{barucci2021deep} the {\em GlyphNet} architecture is introduced, which presents classification results on Hieroglyphs.
In~\cite{resler2021deep} classification is done both by period and by site, utilizing multiple CNNs and inferring with a voting ensemble approach.
Their dataset contains images of archaeological tools and artifacts, mostly found in good preservation conditions.
We compare our results to those of the recent models of~\cite{anwar2021deep,barucci2021deep} and to those of the backbone model of~~\cite{resler2021deep}, which work on 2D data and made their code available.

To generate drawings from 3D archaeological data, in~\cite{kolomenkin2013reconstruction, gilboa2013computer,wilczek2018computer} the sought-after curves are mathematically defined. 
Based on this definition nice results are produced for a few available 3D shapes.
Since 3D archaeological data is even scarcer than 2D data, we are the first to address the problem in 2D.

\vspace{0.01in}
\noindent
{\bf 2D archaeological datasets.}
A dataset of $4,310$ grayscale images of hieroglyphs is presented in \cite{franken2013automatic}.
The artifacts are not well-preserved, however the variety in each class is relatively low. 
A dataset that comprises 
photographs of $6,770$ artifacts of different types
is presented in \cite{resler2021deep}, most are relatively well-preserved, with high variation within each class.
The released version of the dataset is in a much lower resolution than that of the original, therefore we do not experiment on it.
The dataset published in \cite{anwar2021deep} contains $18,225$ images of ancient Roman coins.
The coins are well-preserved and the variety within each class is low.
We present a new dataset, which is not only the first to contain labeled pairs of images and drawings, but also contains challenging artifacts in terms both of preservation and of class variety.

\vspace{0.01in}
\noindent{\bf Semi-supervised multi-modal learning.}
This task aims to integrate information from multiple modalities and learn shared knowledge, making use of both labeled and unlabeled data.
Approaches such as pseudo-labeling~\cite{yoon2022semi, sohn2020fixmatch}, teacher-student distillation~\cite{islam2021dynamic, chen2020big}, or
co-training~\cite{blum1998combining, qiao2018deep, yang2021deep}
do not fully leverage the nature of our paired dataset.
Our work is somewhat related to teacher-student, as we aim to transfer knowledge from one encoder to another.
The difference however is in the quality of the input of the teacher and the student, which use drawings \& images, respectively. 

There are also works which combine sketches and images in various ways~\cite{dey2019doodle,bhunia2021more, zhang2016sketchnet, yang2019person, sain2021stylemeup, sain2023clip, dikkala2022sketching, wang2016deep}.

\section{Method}
\label{sec:method}

Our goal is to design a model which, given only an image of a 3D decorated archaeological artifact, will output an embedding vector representation that can be used for analysis applications, in particular classification or retrieval.
Such images differ from natural images in several manners.
First, they are monochromatic.
Second, the quality of the artifacts is poor, missing some  contours,  while others are created due to noise.
Finally, the photos 
are often  in poor condition.

Sometimes, these images are associated with drawings, created by special archaeological artists, as shown in Figure~\ref{fig:pairs}.
These drawings consist of clear edges and their quality is superior to the quality of the corresponding photographs, for a couple of reasons: 
Due to the artists' vast experience, they can draw parts of the contours, even if they are abraded. 
In addition, the artist may have access to the real artifact when drawing it and can see the 3D features that are unclear in the image.
Hence, these drawings encapsulate  important domain knowledge in them.
However, drawings and images are not necessarily aligned to each other, in terms of the actual geometric alignment of the edges, additional edges, missing edges, and missing damages.
We will show that despite these drawbacks, when paired images and drawings are available during training, image representation is improved, in comparison to learning only from images.

\begin{figure}[t]
\centering
\includegraphics[height=0.5\linewidth, width=1.0\linewidth]{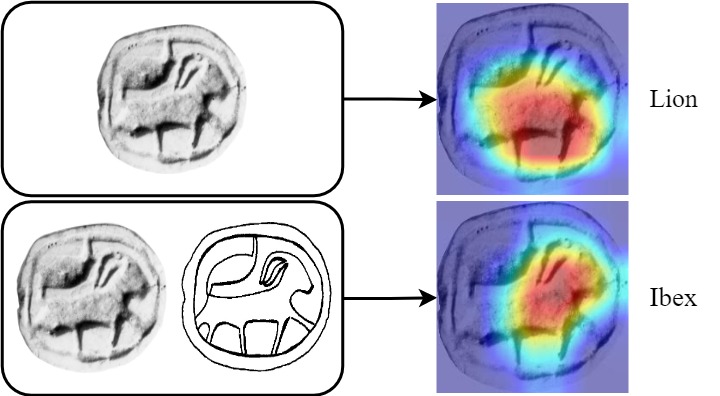}
\caption{{\bf Training with and without drawings.}
When training only using images (top), the model focuses on the torso of the ibex, which leads to misclassification as a lion.
Conversely, thanks to the drawing, our model focuses on the horns and the head (bottom), and classifies the image correctly.
}
\label{fig: gradcam}
\end{figure}

\begin{figure*}[ht!]
\centering
\includegraphics[width=0.9\linewidth, height=0.32\linewidth]{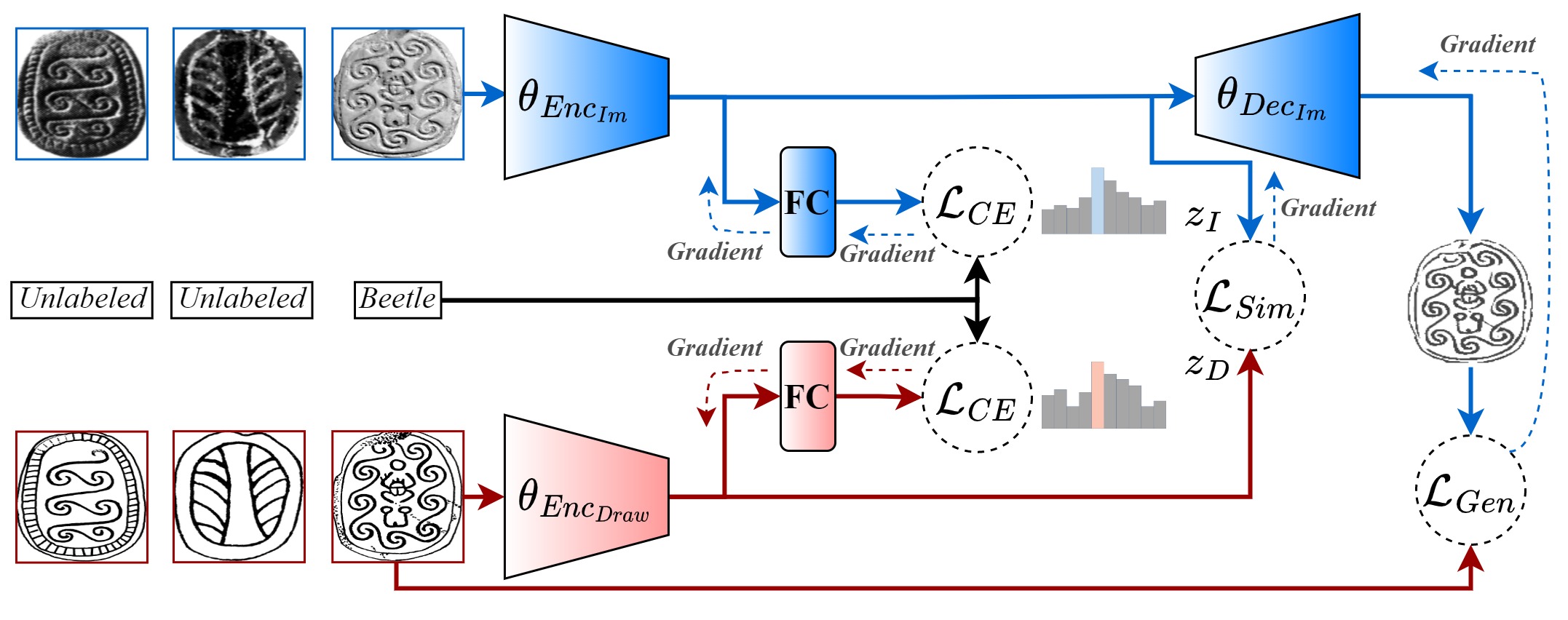}
\caption{
{\bf Model.}
This figure illustrates the processing of a batch of image-drawing pairs ($3$ in this example), where some of them are unlabeled and some are.
The image and its corresponding drawing are encoded, where
\(\theta_{Enc_{Draw}}\) and \(\theta_{Enc_{Im}}\)   represent the parameters of their encoders, respectively.
The FC components represent the classifiers.
The image decoder, whose parameters are \(\theta_{Dec_{Im}}\), generates the reconstructed drawing. 
The loss \(\mathcal{L}\) consists of three components: \(\mathcal{L}_{CE}\) for classification, \(\mathcal{L}_{Gen}\) for image-to-drawing generation
and~\(\mathcal{L}_Sim\), whose goal is to maximize the similarity between pair embeddings.
In case of an unlabeled pair, the classification component is ignored.
Thus, we freeze \(\theta_{Enc_{Draw}}\) and update the other components.
In case of a labeled pair, \(\theta_{Enc_{Draw}}\) is updated due to \(\mathcal{L}_{CE}\); \(\theta_{Enc_{Im}}\) is  updated due to all the components of the loss function; and \(\theta_{Dec_{Im}}\) is updated via \(\mathcal{L}_{Gen}\).
\label{fig: model}}
\end{figure*}

We propose to utilize the drawings during training to improve image representation.
During inference, however, images are the sole input, since in most cases drawings are unavailable.
Intuitively, a drawing of an object can be considered as an augmentation of the photo, one which expresses the object's shape as an edge map and makes it easier to extract features that are difficult to obtain from the image.
This is demonstrated in Figure~\ref{fig: gradcam}, where the drawing enables our model to differentiate between engravings of ibexes and lions.
Using only images, the localization map~\cite{selvaraju2017grad} focuses on a region  (torso) in which the ibex and the lion engravings are  hardly distinguishable, whereas when trained with our method, the model's focus is on the horns, which distinguishes between the two classes.
This intuition is reinforced in our experiments in Section~\ref{sec:ablation}.
Interestingly, since training is performed also for drawings, it enables our model to be used for a generative task---drawing an artifact, i.e. {\em image-to-drawing} in the archaeological domain.

Since in archaeology labeled data is scarce, we propose a training process that utilizes both labeled and unlabeled data, which exists in much larger numbers.
We show that the mere existence of image-drawing pairs, even when unlabeled, helps.
In particular, Section~\ref{sec:ablation} shows that by training in this semi-supervised manner, we achieve better results than by using just the labeled data.
Only classes that are unknown to the model appear in the unlabeled data.

\vspace{0.01in}
\noindent
{\bf The model.}
As shown in Figure~\ref{fig: model}, both labeled and unlabeled pairs of images and drawings  are received as input, randomly in each batch, where most of the pairs are naturally unlabeled.
The key idea is to optimize image embedding for classification, by maximizing the similarity between paired images and drawings.
This is since training a network for drawing classification is easier than for images, as the shape is much clearer in drawings.
Thus, we assume that the feature map of a drawing, represented by an embedding vector, is more informative than that of its corresponding image.
Under this assumption, making the image embedding similar to its paired drawing embedding, contributes significantly to improve image representation. 

Our model, which realizes this idea, consists of two encoders, $Enc_{Draw}$ and $Enc_{Im}$, whose inputs are drawings,~$D$, and the corresponding images, $I$.
Two fully-connected layers of the same structure are trained for classification.
The output of the image encoder is fed into a decoder, $Dec_{Im}$, whose goal is to reconstruct drawings from images.
Let us denote the parameters of the drawing encoder by \(\theta_{Enc_{Draw}}\) and of the image encoder and image-to-drawing decoder by \(\theta_{Enc_{Im}}\) and \(\theta_{Dec_{Im}}\), respectively.

Our method  is general and may train using many types of encoder/decoder backbone architectures.
We will show in Section~\ref{sec:results} that our method significantly improves the results irrespective of the chosen backbone.

\vspace{0.01in}
\noindent
{\bf Losses.}
Two losses use the outputs of the encoders, the {\em similarity loss} (\(\mathcal{L}_{Sim}\))
 whose goal is to maximize the embedding similarity between the drawing and the matching image, and the {\em cross-entropy loss} (\(\mathcal{L}_{CE}\)) whose goal is to minimize classification errors. 
The output of the image decoder, jointly with the original drawing, are used for yet another loss function, the {\em generation loss} (\(\mathcal{L}_{Gen}\)).

In the unsupervised case, when a label is not given, we would still like to train the two encodings to be similar, by modifying the image encoding to be similar to the drawing encoding, which is kept fixed. 
In addition, the reconstructed drawing should be close to the original drawing. Thus, the gradients of $\mathcal{L}_{Sim}$ and $\mathcal{L}_{Gen}$ are used to update \(\theta_{Enc_{Im}}\), the gradient of $\mathcal{L}_{Gen}$ is used to update \(\theta_{Dec_{Im}}\), and \(\theta_{Enc_{Draw}}\) is frozen. 
Freezing \(\theta_{Enc_{Draw}}\) means that drawings affect and improve images embedding, but not the opposite, in a way that might harm drawings embedding.

In the supervised case, the classifier components are also used. 
Thus, in addition to the above,  the  gradients of the two \(\mathcal{L}_{CE}\) losses update their respective encoders. Only in this case \(\theta_{Enc_{Draw}}\) is updated.

Our image encoder  loss is a sum of the three losses:
\begin{equation}
  \mathcal{L} = \gamma_{1} \cdot \mathcal{L}_{Sim} + 
  \gamma_{2} \cdot \mathcal{L}_{CE} + 
  \gamma_{3} \cdot \mathcal{L}_{Gen}.
  \label{eq:stage_1_loss}
\end{equation}
The weights, \(\gamma_{i}\), are hyper-parameters, chosen by trial and error and their values are
\(\gamma_{1}=0.8\), \(\gamma_{2}=0.15\) and \(\gamma_{3}=0.05\).
We hereby elaborate on the losses.

\begin{figure*}[t]
\begin{center}
\begin{tabular}{c c c c c c c c c c c}
 \includegraphics[height=1.1cm]{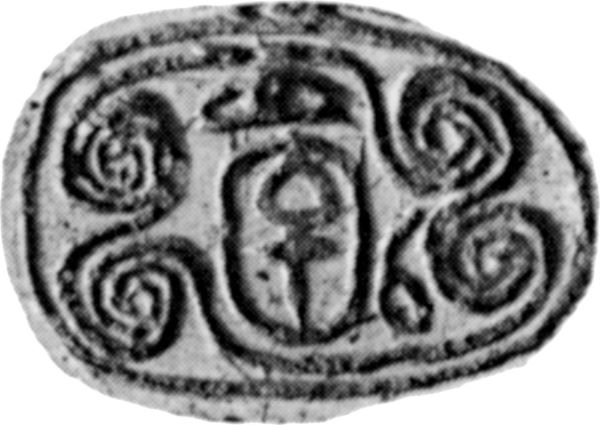} & 
 \includegraphics[height=1.4cm]{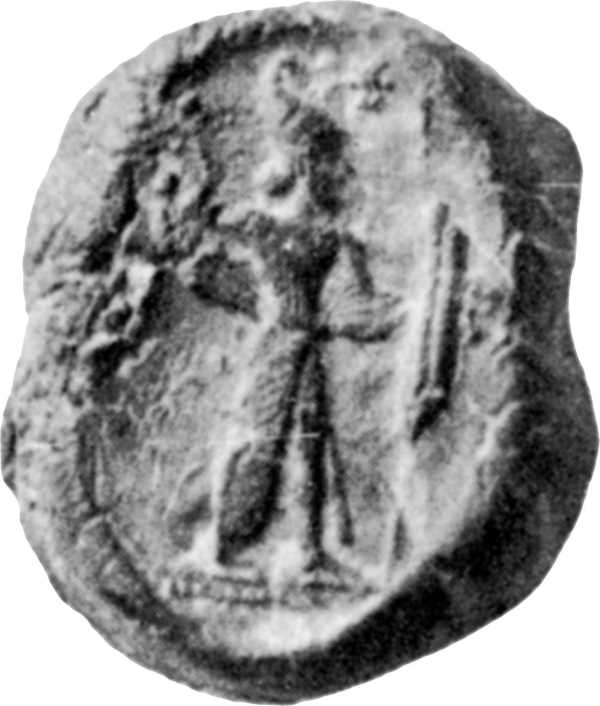} &
 \includegraphics[height=1.4cm]{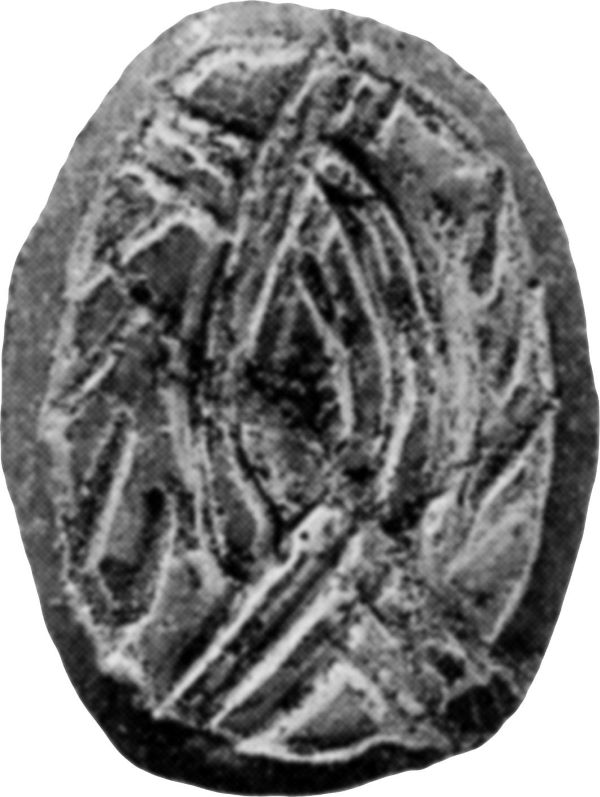} &
 \includegraphics[height=1.4cm]{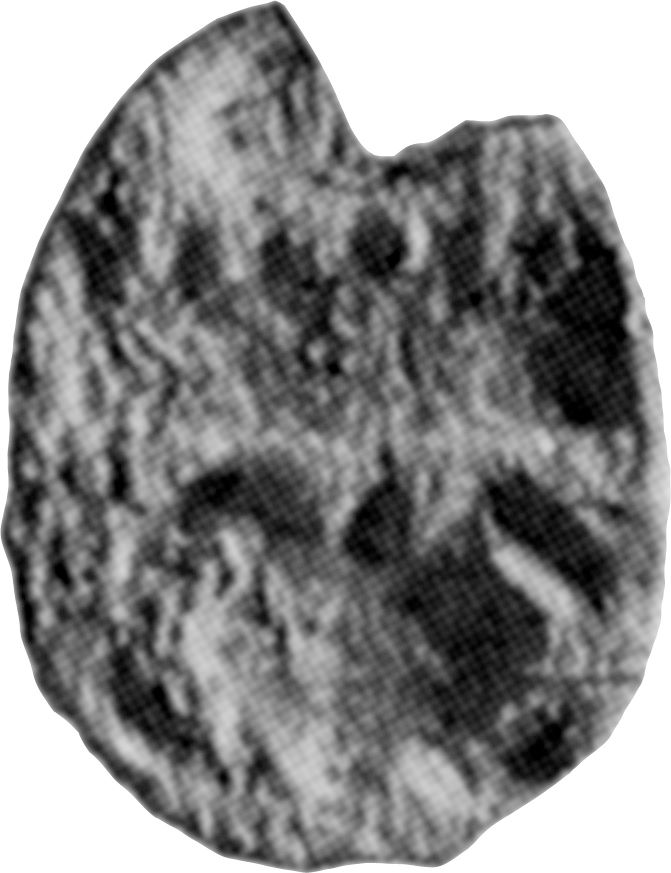} &
 \includegraphics[height=1.4cm]{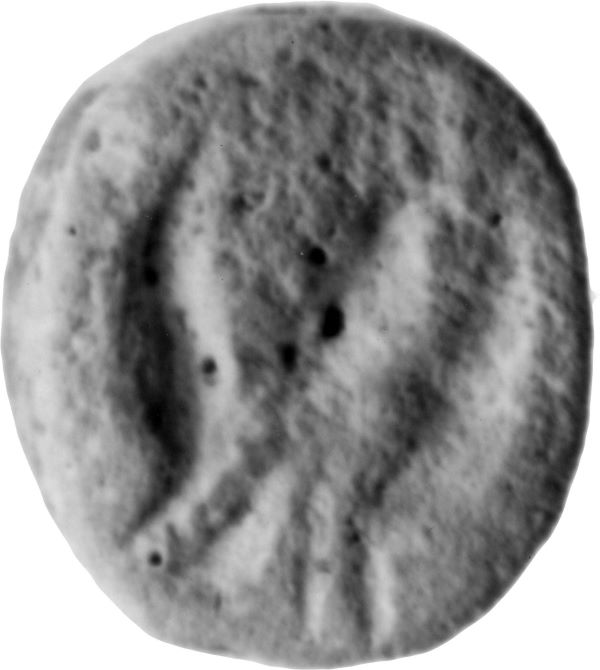} &
 \includegraphics[height=1.4cm]{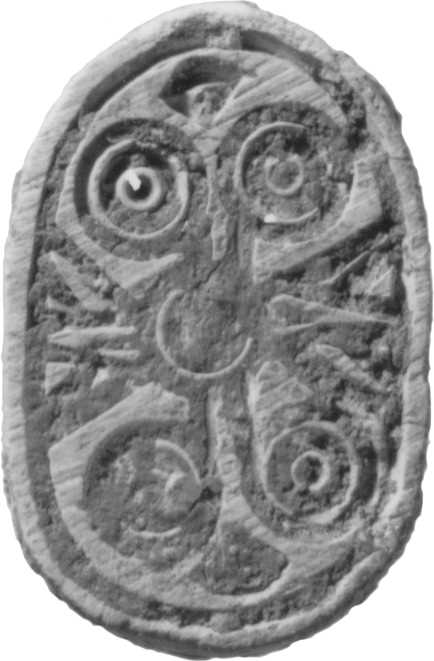} &
    \includegraphics[height=1.4cm]{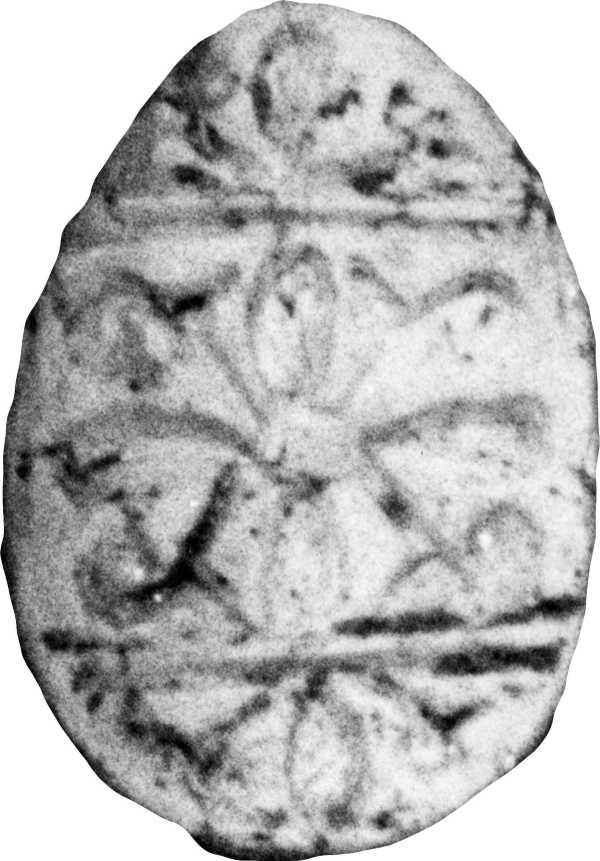} &
 \includegraphics[height=1.1cm]{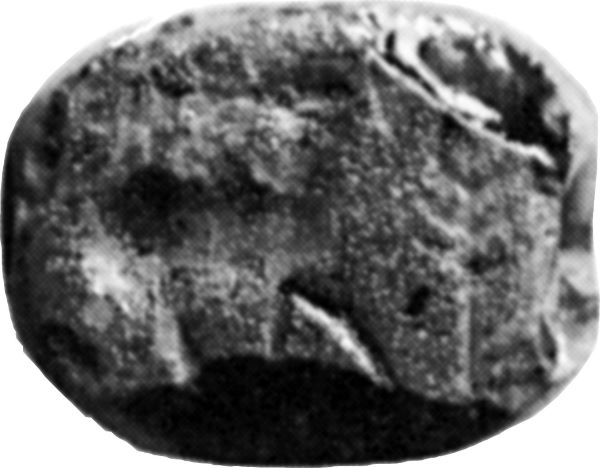} &
 \includegraphics[height=1.1cm]{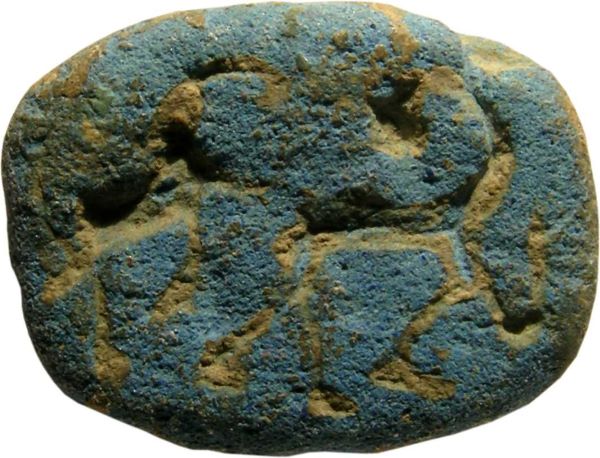} &
 \includegraphics[height=1.1cm]{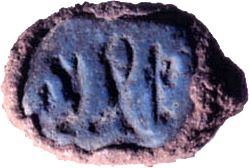} &
 \\  
 \includegraphics[height=1.1cm]{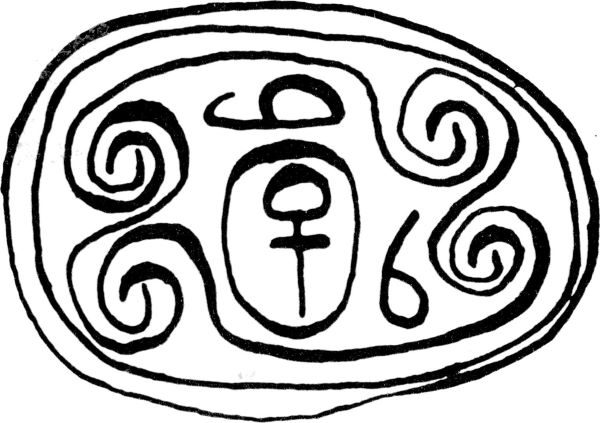} & 
 \includegraphics[height=1.4cm]{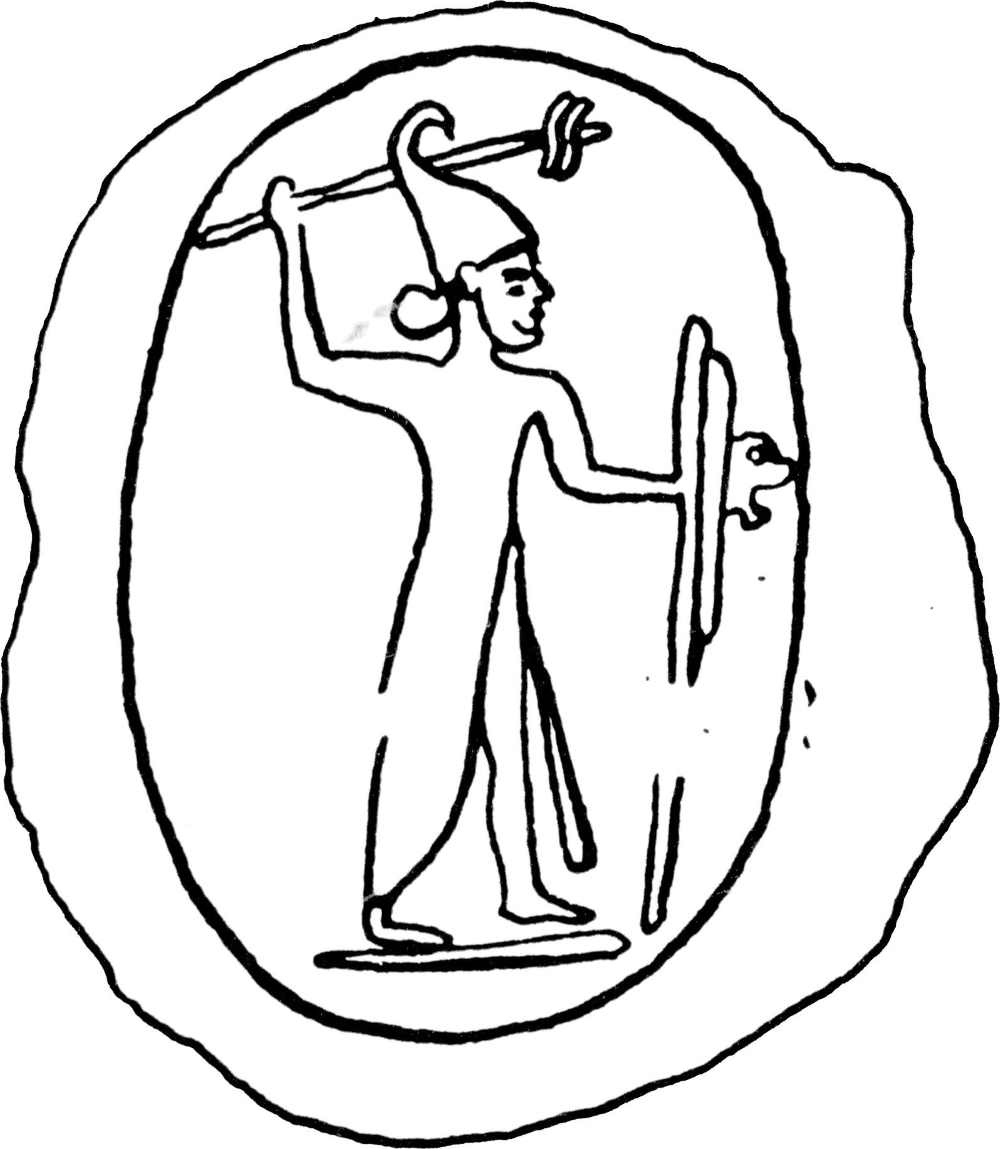} &
 \includegraphics[height=1.4cm]{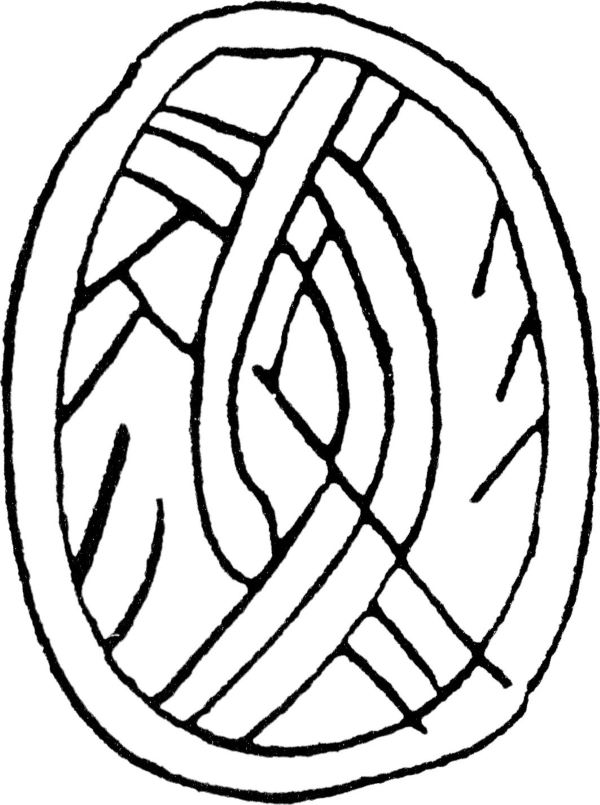} &
 \includegraphics[height=1.4cm]{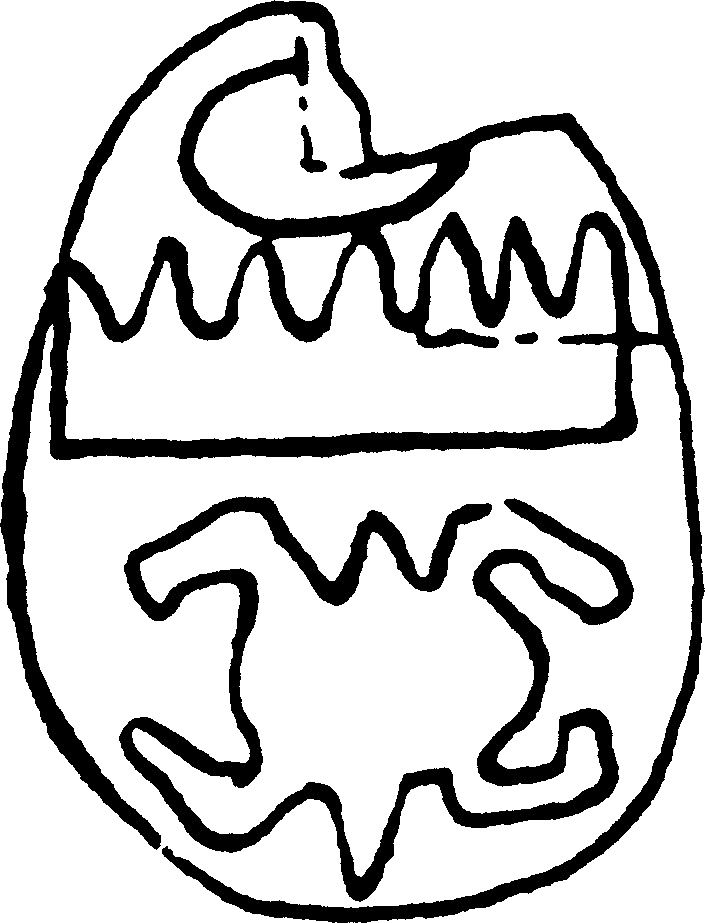} &
 \includegraphics[height=1.4cm]{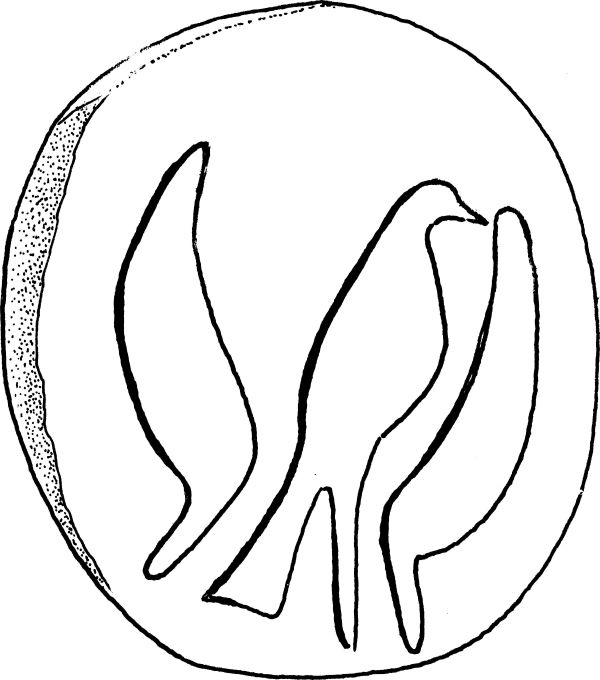} &
 \includegraphics[height=1.4cm]{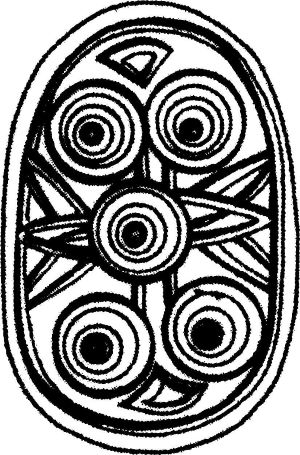} &
   \includegraphics[height=1.4cm]{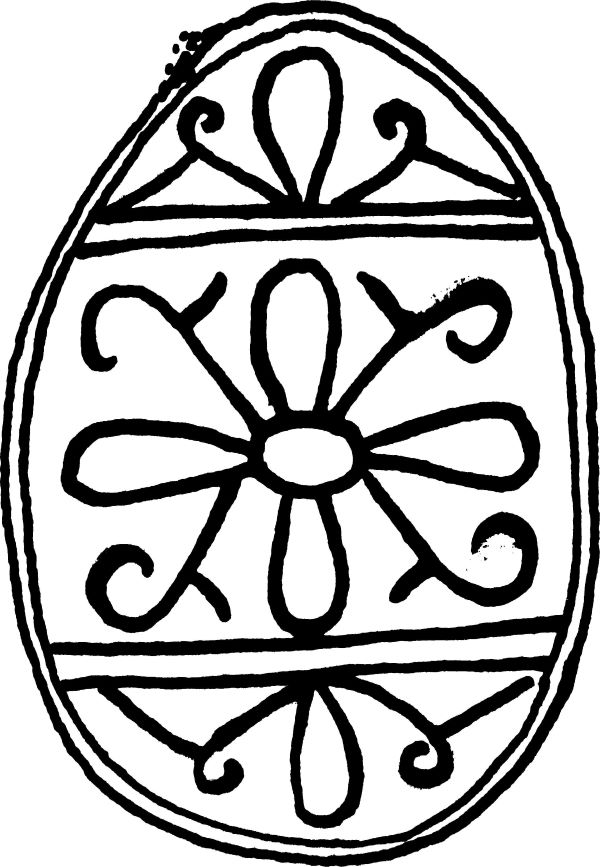} &
 \includegraphics[height=1.1cm]{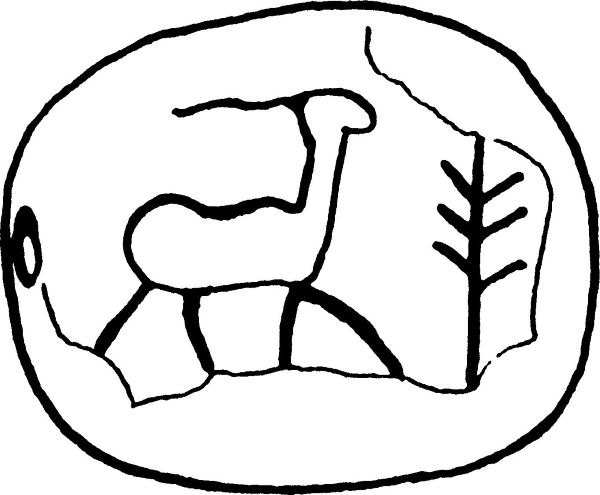} &
 \includegraphics[height=1.1cm]{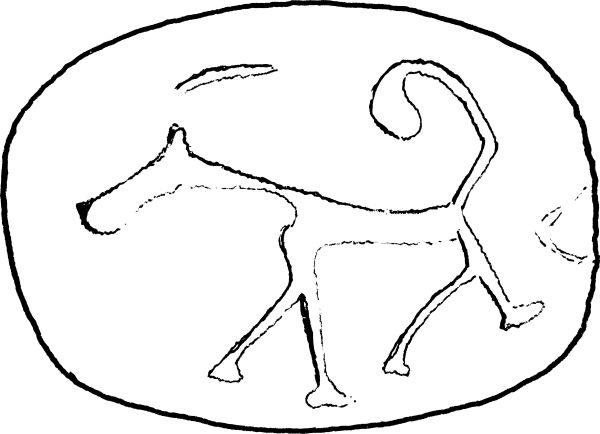} &
 \includegraphics[height=1.1cm]{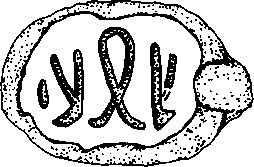} &\\  
\hspace{-0.2in}(a)  Ankh & Anthrop. & Bands & Beetle & Bird & Circles & Cross & Ibex & Lion & Sa \\
\hspace{-0.3in}(b)  $152$ & $130$ & $81$ & $152$ & $95$ & $89$ & $71$ & $101$ & $120$ & $75$ \\
\end{tabular}
\caption{{\bf Shape classes.} (a) shows a single instance of an image-drawing pair for each of the $10$ classes.
(b) is the number of labeled pairs.
}
\label{fig: dataset all classes}
\end{center}
\end{figure*}

To maximize the similarity in the latent space between the image embedding \(z_{I}\) and its corresponding drawing embedding \(z_{D}\), we minimize a negative cosine similarity loss:
\begin{equation}
  \mathcal{L}_{Sim} = -\frac{z_{I}\cdot z_{D}}{\left \| z_{I} \right \|_{2}\cdot\left \| z_{D} \right \|_{2}}.
  \label{eq:cosine-similarity}
\end{equation}
Thus, by maximizing the similarity, we force the image embedding to be closer to the drawings embedding, updating only \(\theta_{Enc_{Im}}\) (and not \(\theta_{Enc_{Draw}}\)).

We train the FC layers for classification with a CE loss, \(\mathcal{L}_{CE}\), one for drawings and the other for images.
The goal is to improve both drawing and image embedding by updating \(\theta_{Enc_{Draw}}\) and \(\theta_{Enc_{Im}}\) for classification.
The CE loss is computed only for the labeled pairs in the batch.

The generation loss takes into account the distance between the reconstructed drawing and the original drawing, using both the \(\mathcal{\ell}_{2}\) norm and a perceptual loss, \(\mathcal{L}_{P}\), 
similarly to~\cite{johnson2016perceptual}.
While \(\mathcal{\ell}_{2}\) aims to minimize mismatches between pixels of a drawing and a generation, the use of the perceptual loss encourages a similar feature representation.
Because the pairs of drawings and images are misaligned, a full pixel-wise match is not possible.
Hence, the combination of \(\mathcal{\ell}_{2}\) and \(\mathcal{L}_{P}\) enables the model to learn a better visually-similar generation.
$\mathcal{L}_{Gen}$ is thus defined as
\begin{equation}
  \mathcal{L}_{Gen} = \alpha \cdot \left \| \tilde{D} - D \right \|_{2} + \beta \cdot \mathcal{L}_{P}.
  \label{eq:generation-loss}
\end{equation}
Here, \(\tilde{D}\) is the reconstructed drawing, $D$ is the original drawing, and \(\alpha\) \& \(\beta\) are weights.
In our implementation, $\alpha=0.3$ and $\beta=0.7$; they were chosen via grid search.

\section{Our new dataset - CSSL}
\label{sec:dataset}

A major contribution of our paper is a novel dataset of pairs of images \& drawings of ancient Egyptian scarabs, called the {\em Corpus of the Stamp-seals of the Southern Levant (CSSL)}.
This is the first dataset that contains paired images and drawings of any class of archaeological artifacts.
The data was collected by seven different archaeologists for their own archaeological research.
Thus, images might be centered and aligned differently between the archaeologists.
Each artifact was classified by an expert archaeologist and was drawn by a trained drafts person. 
Despite the archaeological significance of these findings to their owners, we managed to get permission from all the involved parties to make this data available to the computer vision community.

CSSL contains $6,636$ pairs, out of  which $1,020$ pairs are classified into $10$ classes of scarab shapes and $5,616$ pairs are unclassified.
The classes and the number of objects per class are shown in Figure~\ref{fig: dataset all classes}.
The supplemental material contains additional images of the various classes.

\begin{figure}[t]
\begin{center}
\begin{tabular}{c c c c c}
 \includegraphics[width=0.17\linewidth]{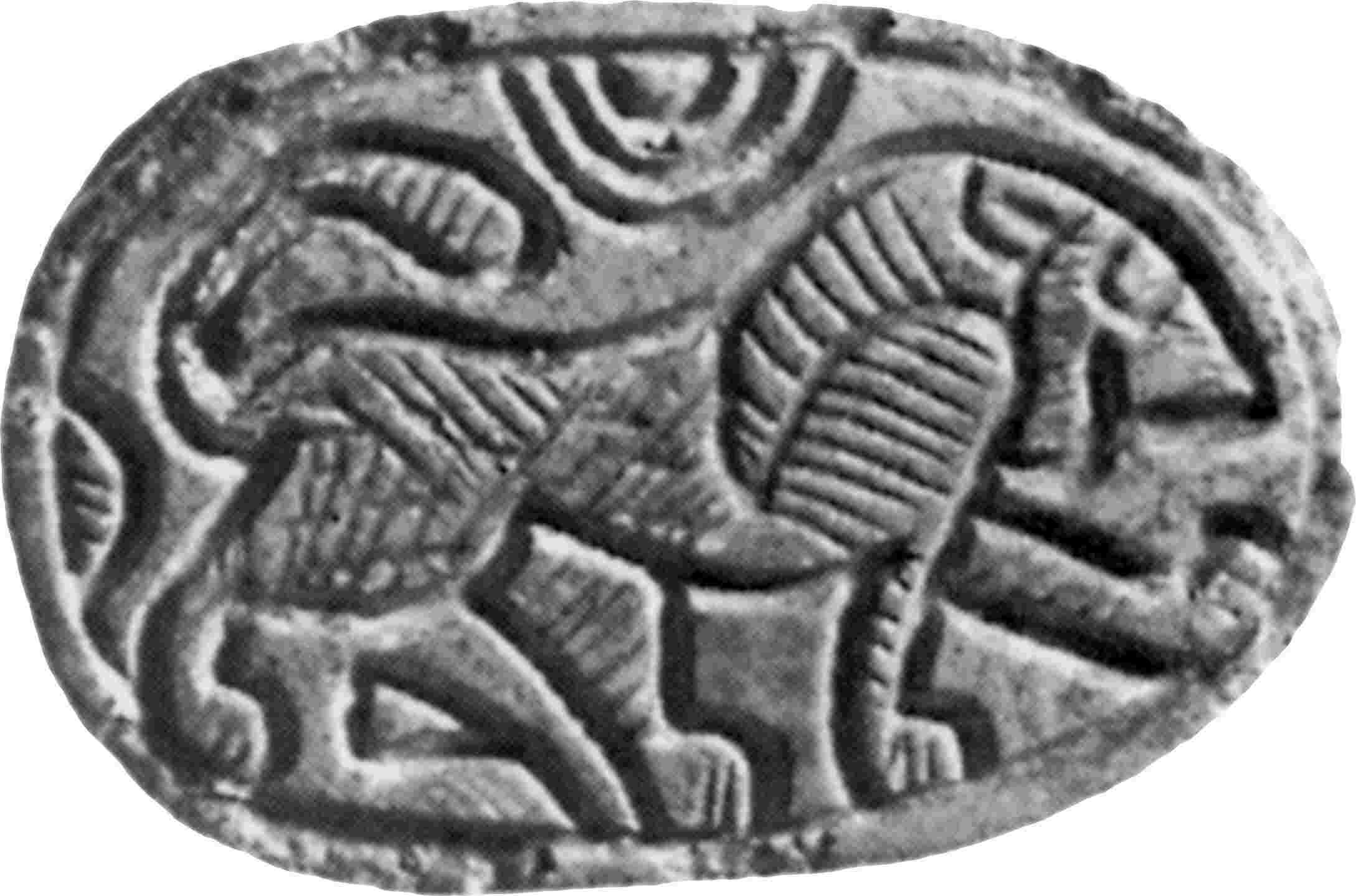} & \includegraphics[width=0.16\linewidth]{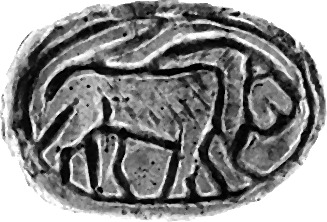} & 
 \includegraphics[width=0.15\linewidth]{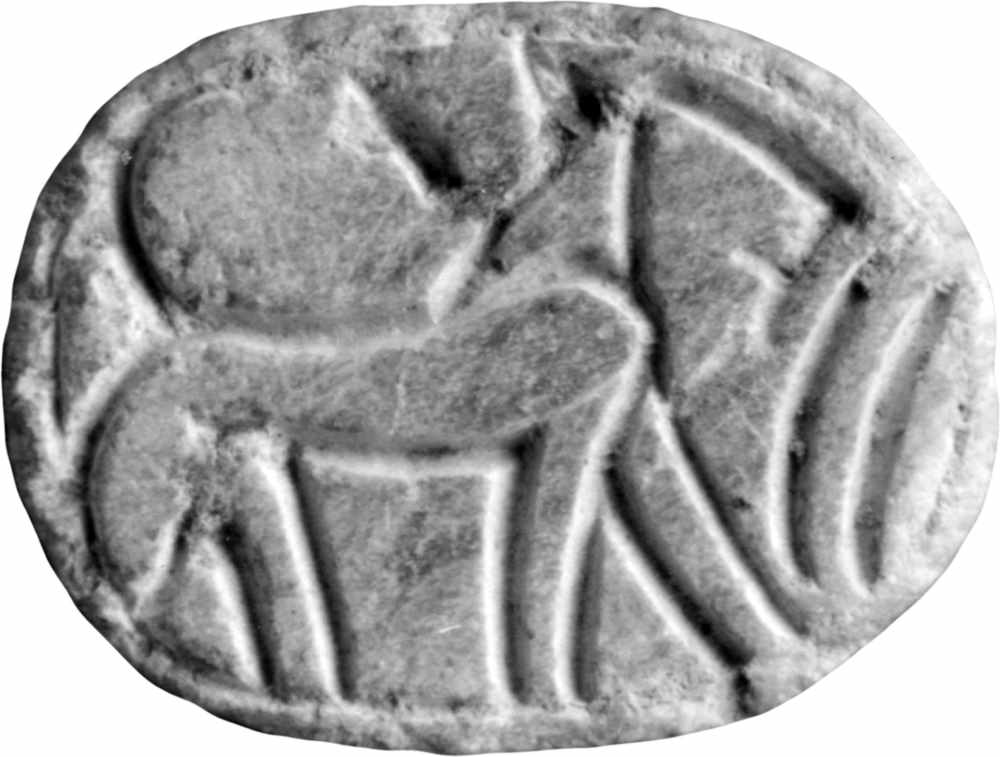} & 
  \includegraphics[width=0.15\linewidth]{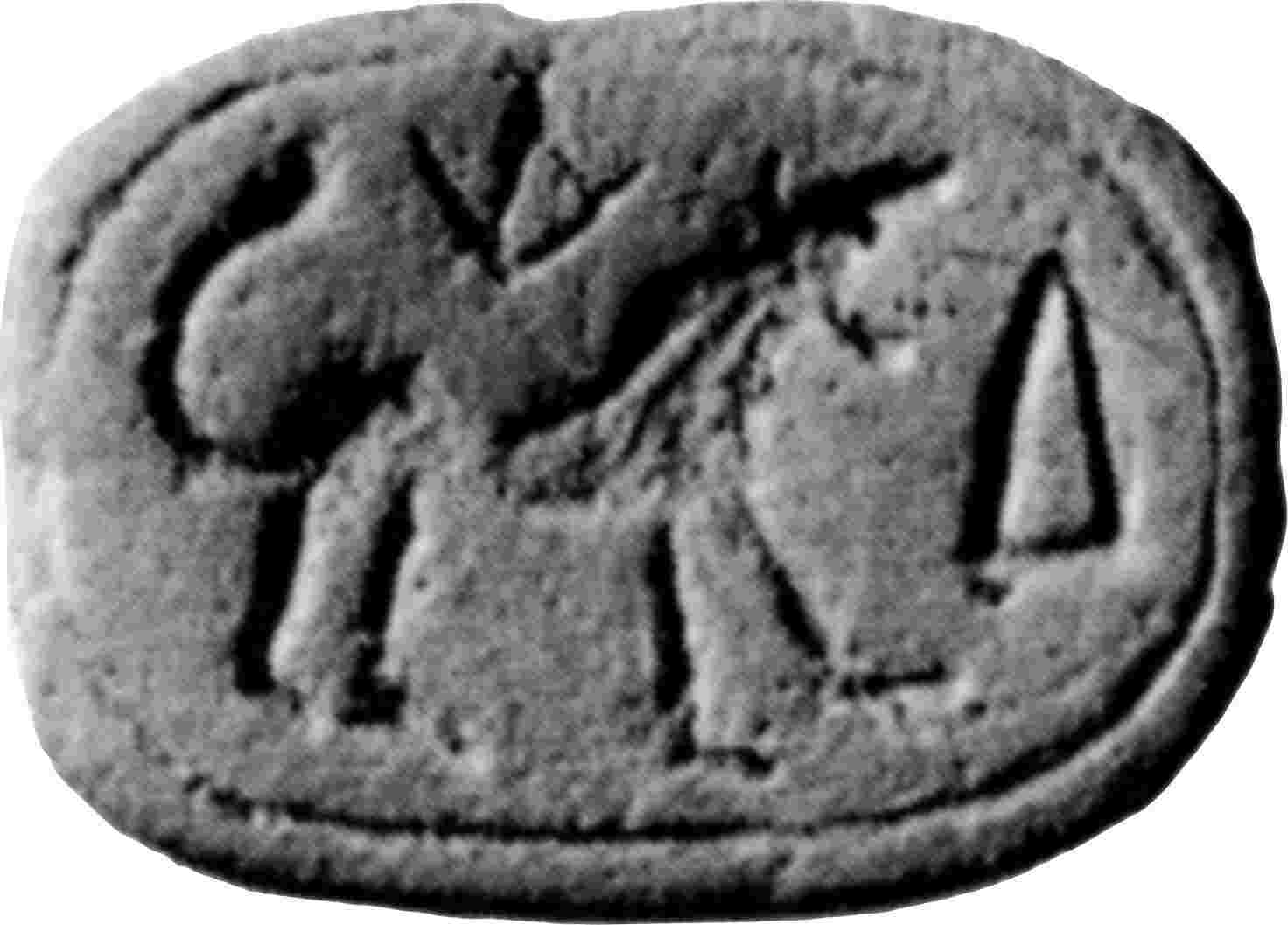} & 
 \includegraphics[width=0.15\linewidth]{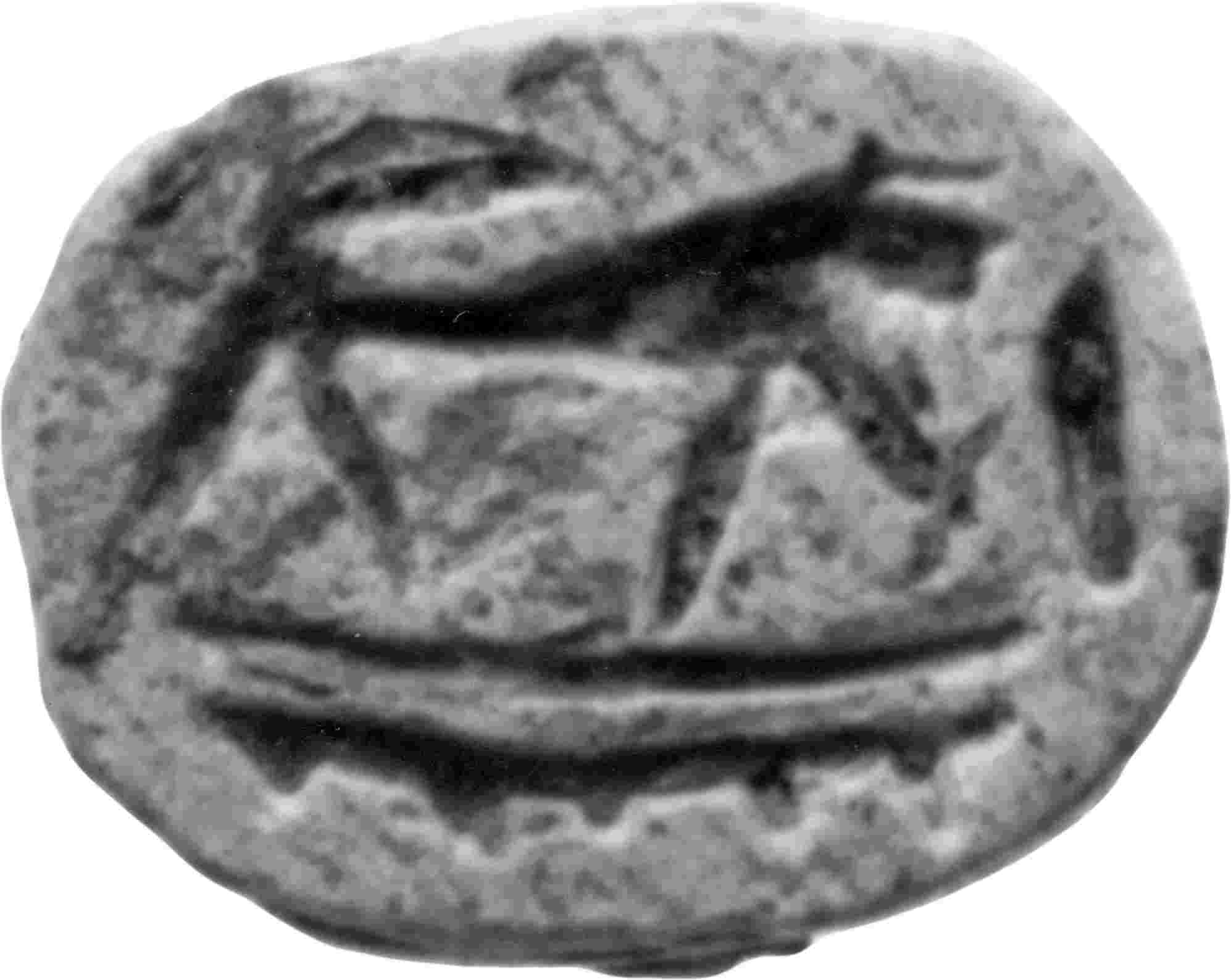} \\  
 \includegraphics[width=0.17\linewidth]{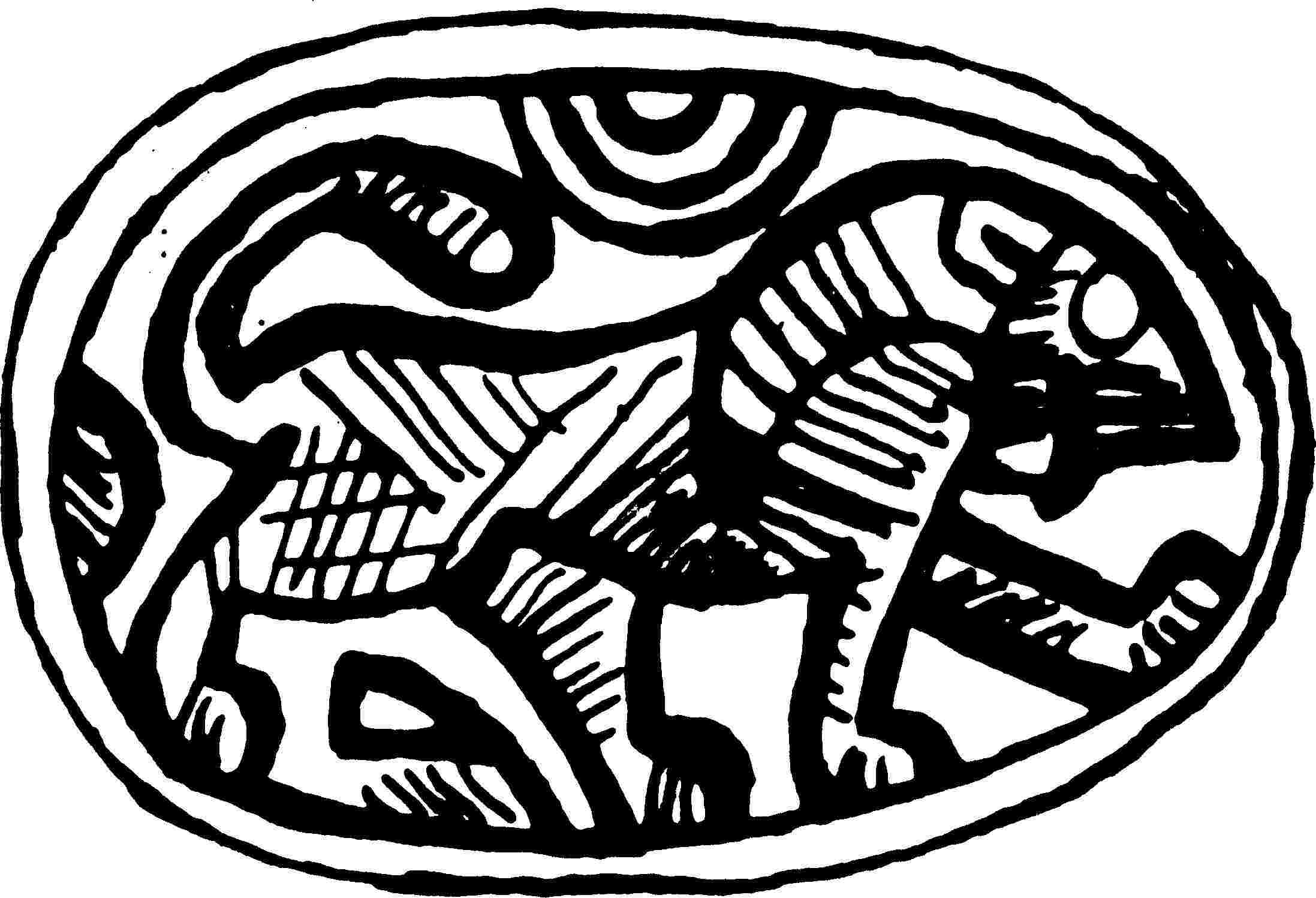} & \includegraphics[width=0.16\linewidth]{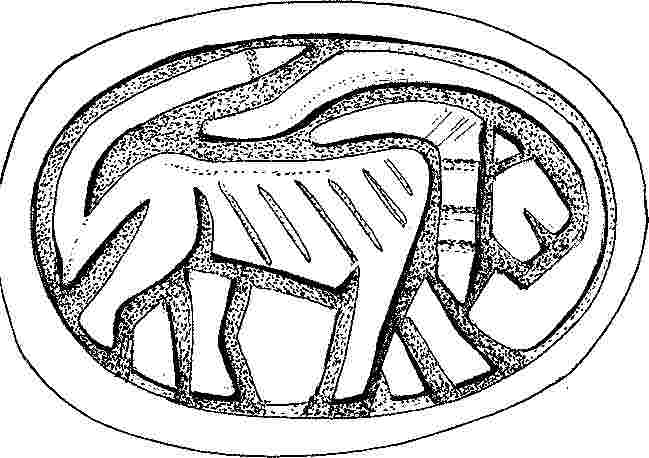} &  \includegraphics[width=0.15\linewidth]{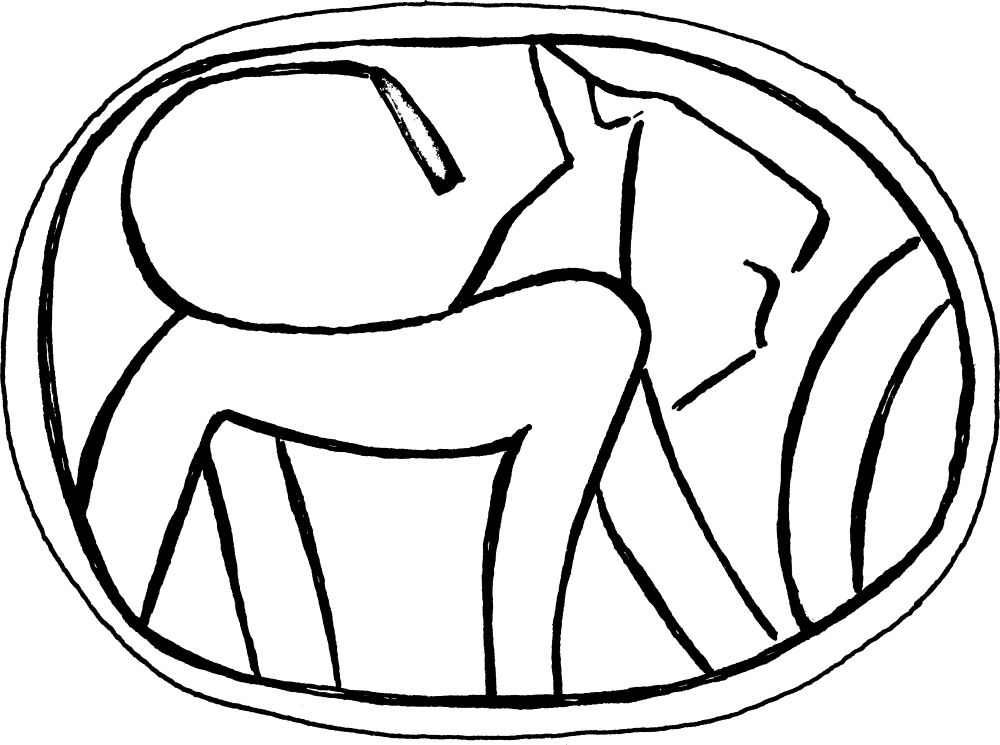} & 
 \includegraphics[width=0.15\linewidth]{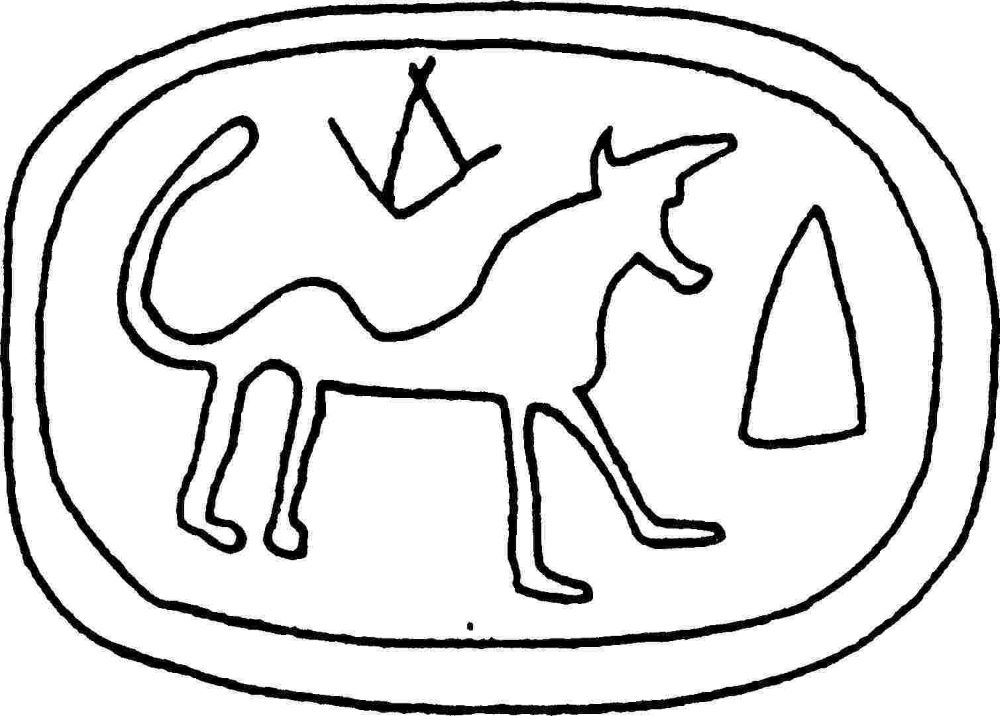} & 
 \includegraphics[width=0.15\linewidth]{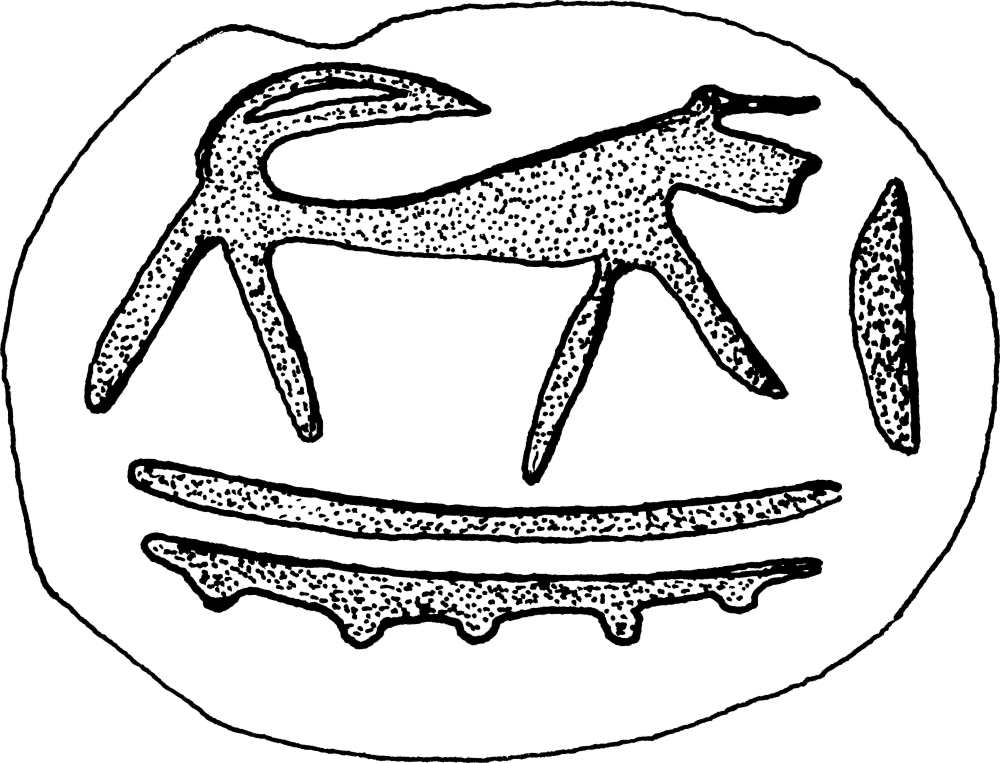} \\  
\hspace{-0.1in}(a) M.B.  & M.B. & L.B. & Iron & Iron \\
\hspace{-0.35in}(b) Early & Late & & Early & Late \\
\hspace{-0.45in}(c) 144 & 374 & 217 & 43 & 44
 
\end{tabular}
\caption{{\bf Sub-period classes.} 
All the findings are decorated by lions. 
They differ in shape and are dated to different periods~(a) and sub-periods~(b).
The bottom row~(c) shows the number of labeled pairs in each sub-period (not just for lions).}
\label{fig: sub periods}
\end{center}
\end{figure}

This dataset has a secondary classification into three periods and five sub-periods.
In particular, $955$ pairs are labeled into Middle Bronze (MB), Late Bronze (LB) and Iron ages and $296$ of the labeled data do not have a period label.
The Middle Bronze age and the Iron age are divided into two  sub-periods each.
Thus, $820$ pairs are labeled into sub-periods.
Figure~\ref{fig: sub periods} illustrates the difficulty of classification into periods, as lion-decorated scarabs, which are relatively similar to a non-expert, are related to different periods.

Figures~\ref{fig: dataset all classes}-\ref{fig: sub periods} demonstrate why Archaeological datasets might be very challenging for  computer vision techniques. 
The artifacts are eroded, causing the shapes to be hard to discern. 
In addition, the artifacts were made over a period of hundreds of years, which naturally resulted in the shapes of the objects from the same class to vary significantly.

\section{Experimental results}
\label{sec:results}


\noindent
{\bf Experimental setup.}
We evaluated our results using a variety of backbones as encoders:
Resnet101~\cite{he2016deep}, DenseNet161~\cite{huang2017densely}, EfficientNetB3~\cite{tan2019efficientnet} (used in~\cite{resler2021deep}), CoinNet~\cite{anwar2021deep}, pretrained on ImageNet~\cite{deng2009imagenet}, and
Glyphnet~\cite{barucci2021deep, glyphnetpytorch2021alekseev}. 
The latter two are designed for archaeological data.
For each encoder, we implemented a decoder, such that the setup is similar to that of UNet~\cite{ronneberger2015u}.
We show that for each backbone  and  task, our method significantly improves the results, demonstrating its generality.

We present the average result obtained for a two-fold cross-validation process, 50\% train and 50\% test.
We use the accuracy measure for classification and \emph{mean average precision (mAP)} for retrieval, both are the most-commonly used measures.
For mAP, we evaluate P@1 and P@10.
Additional metrics and Confusion matrices, are given in the supplemental material.

\vspace{0.01in}
\noindent
{\bf Classification and retrieval.}
Table~\ref{table: classification and retrieval by shape on our dataset} shows that our method indeed improves the shape classification accuracy of each of the five models on CSSL by $7.1\%$-$18.3\%$, and the mAP score by $0.08$-$0.23$.

\begin{table}[t]
\centering
\begin{tabular}{ |l|c|c|c|  }
 \hline
 Model & Class. & P@1 & P@10\\
 \hline
 \hline
  DenseNet161~\cite{huang2017densely}   & 80.8\%    &0.78 &0.77\\
 DenseNet161~\cite{huang2017densely}+ours   & 90.5\%    &{\bf0.90} &{\bf0.89} \\
 \hline
 Resnet101~\cite{he2016deep}   & 82.5\%    &0.78  &0.77\\
 Resnet101~\cite{he2016deep}+ours   & 89.6\% & 0.86 & 0.86\\
 \hline
 EfficientNetB3~\cite{tan2019efficientnet}   & 72.9\%    &0.64  &0.61\\
 EfficientNetB3~\cite{tan2019efficientnet}+ours   & 87\%    &0.85  &0.84\\
 \hline
 \multicolumn{4}{c}{Architectures for Archaeology}
  \\
 \hline
  CoinNet\cite{anwar2021deep}   & 79.6\%    & 0.77 & 0.76 \\
 CoinNet\cite{anwar2021deep}+ours   & {\bf90.8\%}   & 0.88 & 0.88 \\
 \hline
 Glyphnet\cite{barucci2021deep}   & 55.4\%    & 0.43 & 0.43\\
 Glyphnet\cite{barucci2021deep}+ours   & 73.7\%    & 0.65 & 0.64\\
 \hline
\end{tabular}
\caption{{\bf Shape classification and retrieval on our dataset.} 
Our method is general and can utilize a variety of backbones.
It outperforms all previous methods when compared on the same backbone.
The best classification and retrieval results are obtained when using our method on top of DenseNet161 and CoinNet respectively.}
\label{table: classification and retrieval by shape on our dataset}
\end{table}

Table~\ref{table: classification by period on our dataset} shows the results of classification by period.
We trained our model first on shape classification and then
fine-tuned it for periods.
For each backbone, our method improves the results of $3$-period classification by $0.9\%$-$5.1\%$ and of $5$-period by $1.4\%$-$9.0\%$.
The best results are obtained when using our method on top of CoinNet for $3$-period ($84.5\%$) and DenseNet161 for $5$-period (72.7\%).

\begin{table}[t]
\centering
\begin{tabular}{ |l|c|c|  }
 \hline
 Model & 3 Periods & 5 Periods\\
 \hline
  \hline
 DenseNet161~\cite{huang2017densely}   & 81.3\%    &71.3\%\\
 DenseNet161~\cite{huang2017densely}+ours   & 84.0\%    &{\bf72.7\%}\\
 \hline
 Resnet101~\cite{he2016deep}   &  81.1\%   &68.9\%\\
 Resnet101~\cite{he2016deep}+ours   & 83.8\%  & 72.4\%\\
 \hline
 EfficientNetB3~\cite{tan2019efficientnet}   & 77.3\%    &66.6\%\\
 EfficientNetB3~\cite{tan2019efficientnet}+ours   & 82.4\%    &71.6\%\\
 \hline
 \multicolumn{3}{c}{Architectures for Archaeology}\\
 \hline
  CoinNet\cite{anwar2021deep}   & 83.6\%    & 70.2\%\\
 CoinNet\cite{anwar2021deep}+ours   &  {\bf84.5\%}   & 72.2\% \\
 \hline
 Glyphnet\cite{barucci2021deep}   & 72.2\%    &56.6\%\\
 Glyphnet\cite{barucci2021deep}+ours   & 75.6\%    &65.6\%\\
 \hline
\end{tabular}
\caption{{\bf Period classification on our dataset.} 
Our method improves all models. 
Best results are obtained when using our method on top of DenseNet for $3$-period and CoinNet for $5$-period.
}
\label{table: classification by period on our dataset}
\end{table}

Next, we experimented with the hieroglyphs dataset of~\cite{franken2013automatic}.
The artifacts are not well-preserved, similarly to our dataset.
Thus, we assumed that paired drawings (during training) could improve classification.
However, this dataset does not contain drawings.
Instead, we utilized general illustrations of the hieroglyph types from ~\cite{jsesh2022rosmord, Fabricius2022Googleartsculture, MiddleEgyptianDictionaryWebsite2021fayrose}.
During training we paired each image to a random sample of hieroglyph drawing of the same type; they differ in style and small features.
We used $2$-$7$ drawings for each class.
In contrast to our dataset, here all pairs are labeled.

We compare our results to those of \cite{barucci2021deep}.
They use a train, test and validation split of $70\%$, $15\%$ and $15\%$ respectively.
Most of the dataset is released, but not all, thus we re-split the released subset with the same ratios. 
Table~\ref{table: 70/15/15 classification on glyphs} compares the results presented in \cite{barucci2021deep} on the full dataset to the results attained when training it on the released sub-set, with and without our model.
It is shown that our results outperform~\cite{barucci2021deep}'s.
Thus, we show that even though the drawings are not accurate matches to the images, using them during training improves the classification results.
We note that the diversity within each class in this dataset is low, which makes classification easier than it is for our dataset.

\begin{table}[t]
\centering
\begin{tabular}{ |l|c|c|  }
 \hline
 Model & Dataset & Classification\\
 \hline
  \hline
 Glyphnet\cite{barucci2021deep} & Full \cite{barucci2021deep} & 97.6\% \\
  \hline
 Glyphnet\cite{barucci2021deep} & Released \cite{franken2013automatic} & 99.2\% \\
  \hline
 Glyphnet\cite{barucci2021deep}+ours & Released \cite{franken2013automatic} & {\bf99.4\%}\\
  \hline
\end{tabular}
\caption{{\bf Hieroglyphs classification on \cite{franken2013automatic}.}
Our results are better than those reported in~\cite{barucci2021deep} on the full dataset and those attained on the released subset, when using the same backbone.}
\label{table: 70/15/15 classification on glyphs}
\end{table}

Since this dataset is relatively easy, to further evaluate the benefit of our model, we applied it to a small training set.
We split the released dataset into $20\%$ train and $80\%$ test.
Table~\ref{table: classification on glyphs} compares the results on several backbones.
The classification accuracy improves by $0.5\%$ by our method, which is quite significant for this specific dataset.

\begin{table}[t]
\centering
\begin{tabular}{ |l|c|  }
 \hline
 Model & Classification\\
 \hline
  \hline
 Resnet50~\cite{he2016deep} & 94.5\%    \\
 Resnet50~\cite{he2016deep}+Ours &95.0\% \\
  \hline
 Densnet161~\cite{huang2017densely} & 96.0\% \\
 Densnet161~\cite{huang2017densely}+Ours   &96.5\%\\
  \hline
  \hline
 Glyphnet\cite{barucci2021deep} & 96.2\% \\
 Glyphnet\cite{barucci2021deep}+Ours   & {\bf96.7\%}\\
  \hline
\end{tabular}
\caption{{\bf Hieroglyphs classification on \cite{franken2013automatic}.} 
Our method outperforms all  backbones. 
All the backbones are trained from scratch; similar results are obtained for pre-trained backbones.}
\label{table: classification on glyphs}
\end{table}

\vspace{0.01in}
\noindent
{\bf Image-to-drawing generation.}
We strive to create an informative drawing from a given image of an archaeological artifact.
This task differs from that of classical edge detection for a couple of reasons.
First, unlike natural images, artifacts might be eroded and highly noisy, and edges are expected to be also in the eroded parts.
Second, the drawings are not fully aligned with the image. Recall that the second challenge is handled by  \(\mathcal{L}_{P}\), which aims to maximize similar feature maps, rather than full pixel-wise match.

\begin{figure}[t]
\begin{center}
\begin{tabular}{c c c c}
 \includegraphics[width=1.0cm]{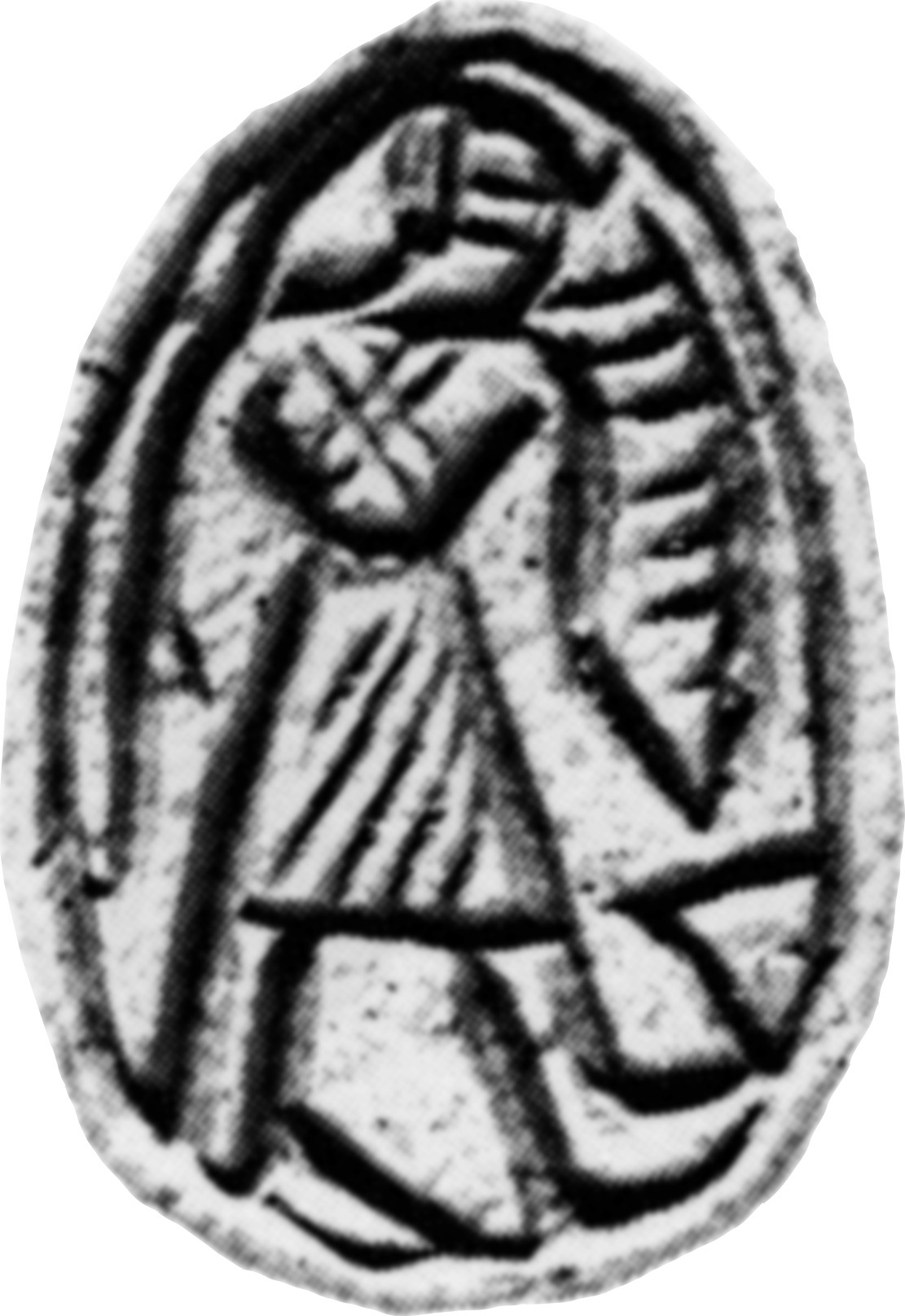} & \includegraphics[width=1.0cm]{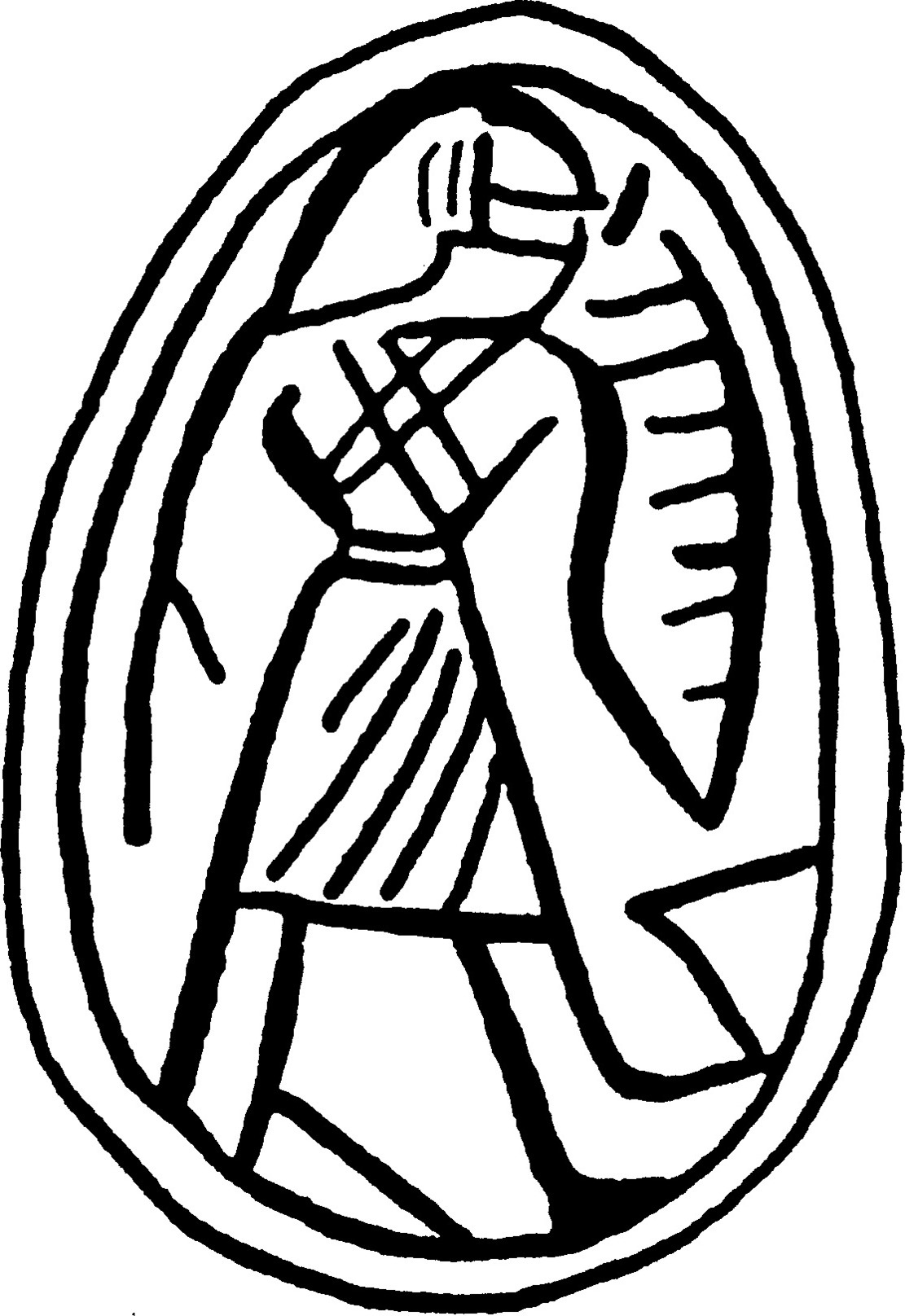} & 
  \includegraphics[width=1.05cm]{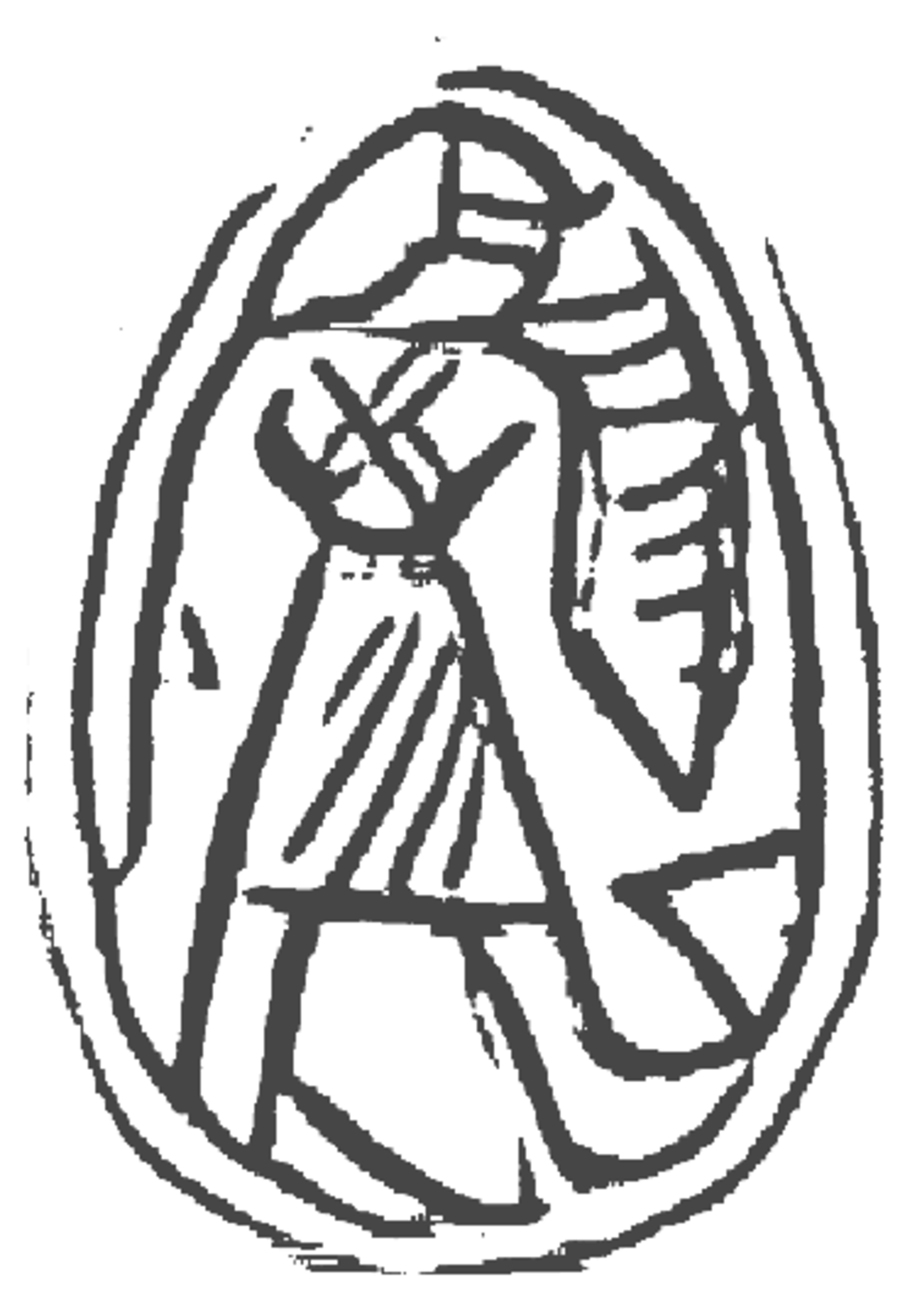} &
 \includegraphics[width=1.0cm]{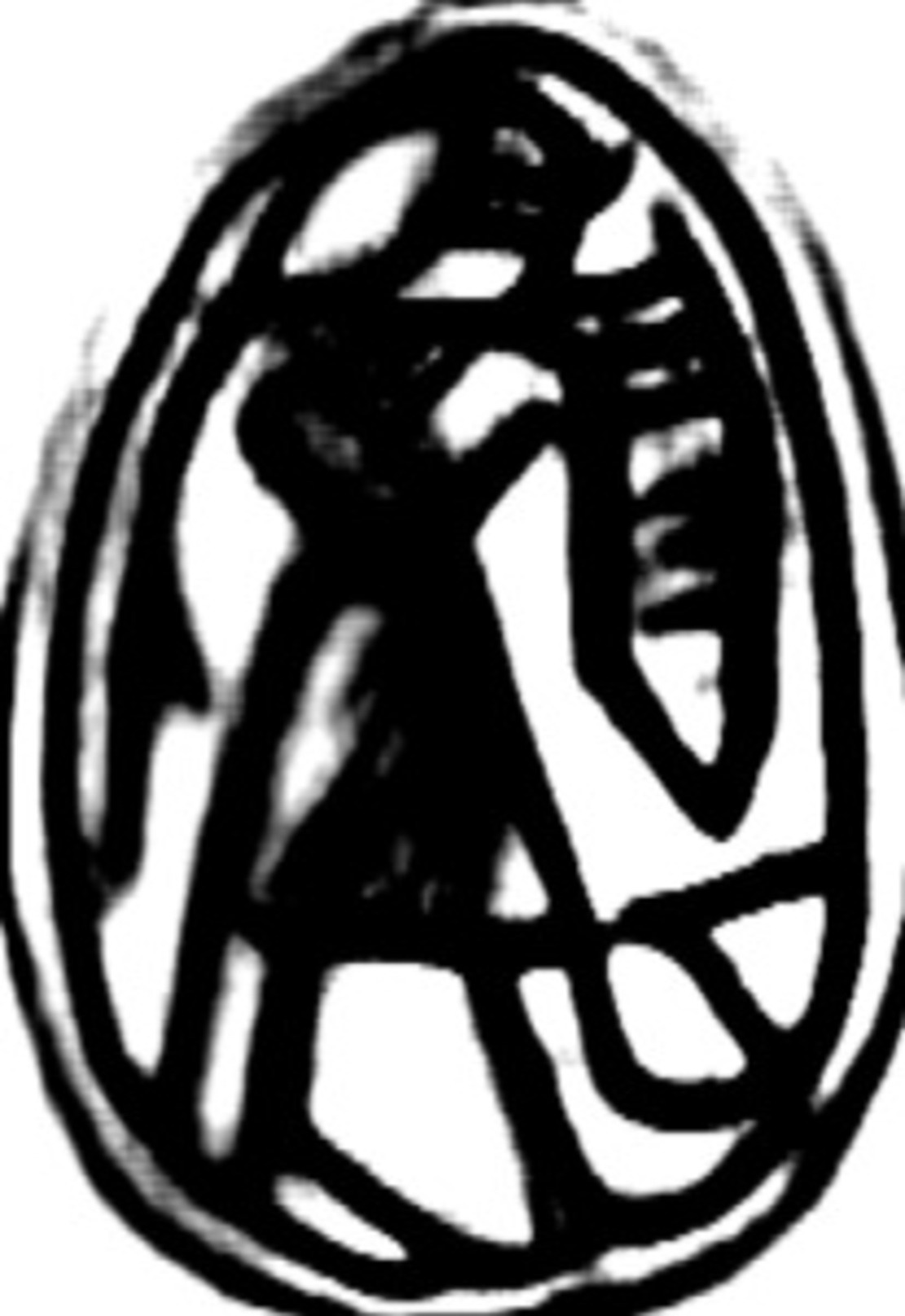} \\ 
 \multicolumn{4}{c}{anthropomorphic}\\[0.1in]
 \includegraphics[height=1.15cm, width=1.3cm]{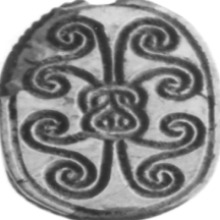} & \includegraphics[height=1.15cm, width=1.3cm]{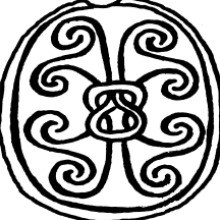} & 
  \includegraphics[height=1.15cm, width=1.3cm]{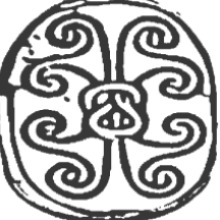} &
 \includegraphics[height=1.15cm, width=1.3cm]{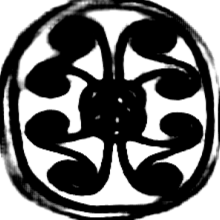} \\ 
 \multicolumn{4}{c}{cross}\\[0.1in]
 \includegraphics[width=1.5cm, height=1.1cm]{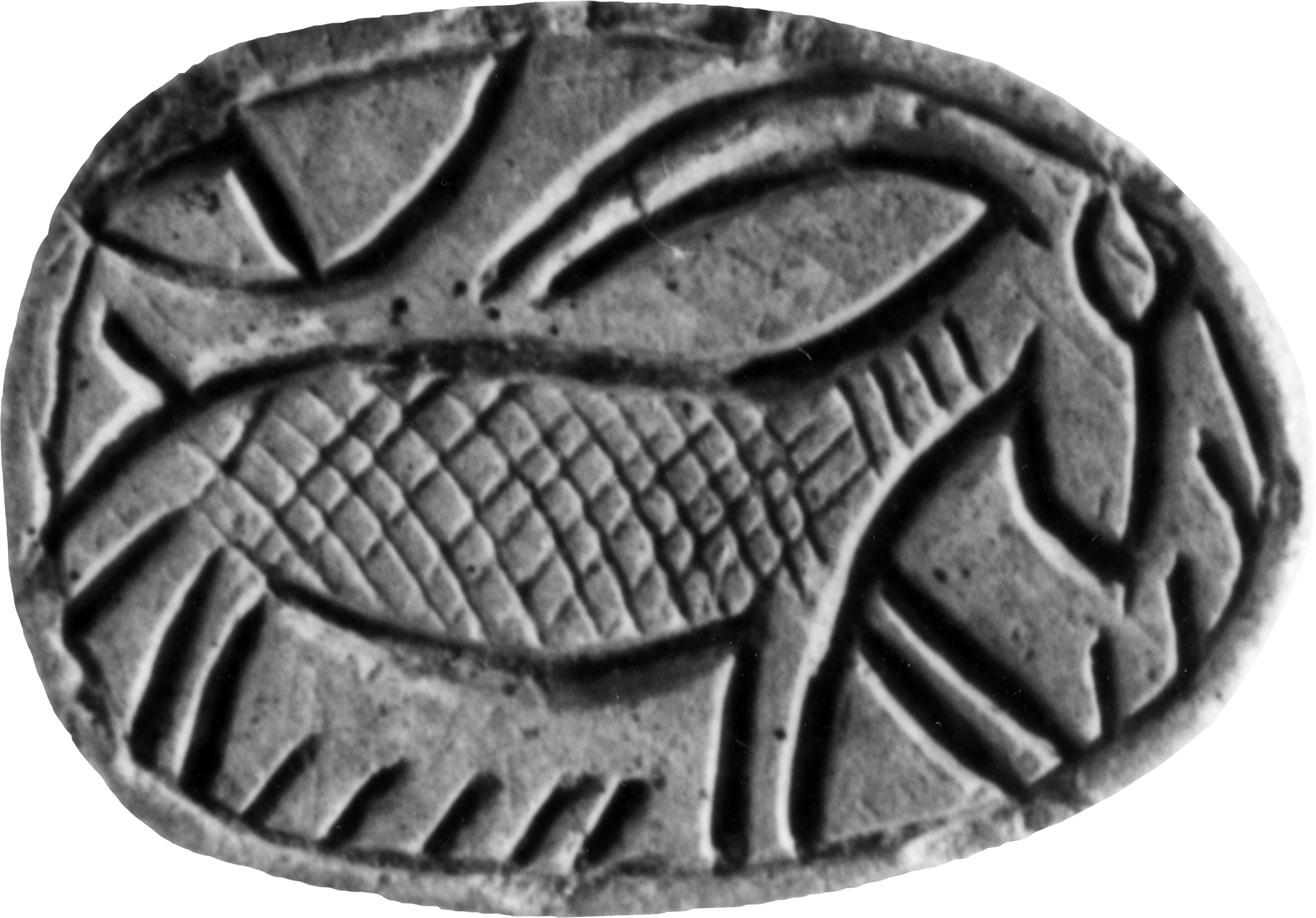} & \includegraphics[width=1.5cm, height=1.1cm]{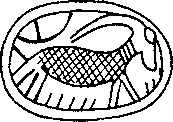} & 
  \includegraphics[width=1.5cm, height=1.1cm]{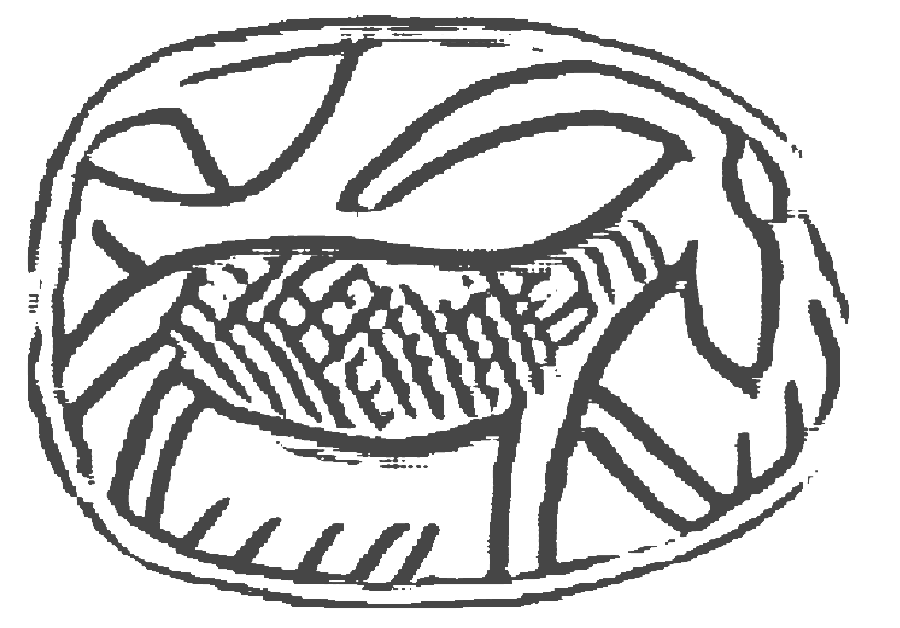} &
 \includegraphics[width=1.6cm, height=1.0cm]{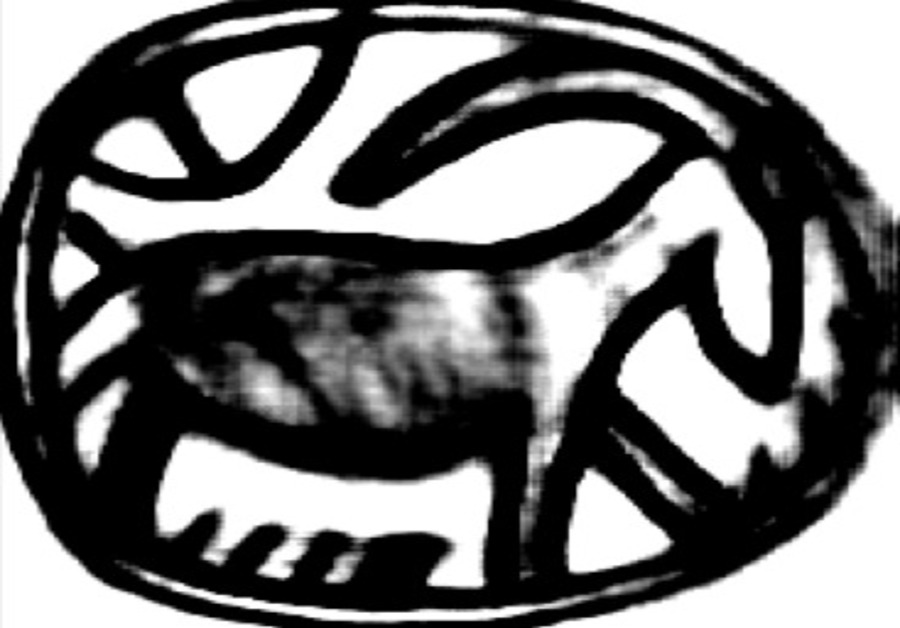} \\
 \multicolumn{4}{c}{ibex}\\[0.1in]
 \includegraphics[width=1.2cm, height=1.4cm]{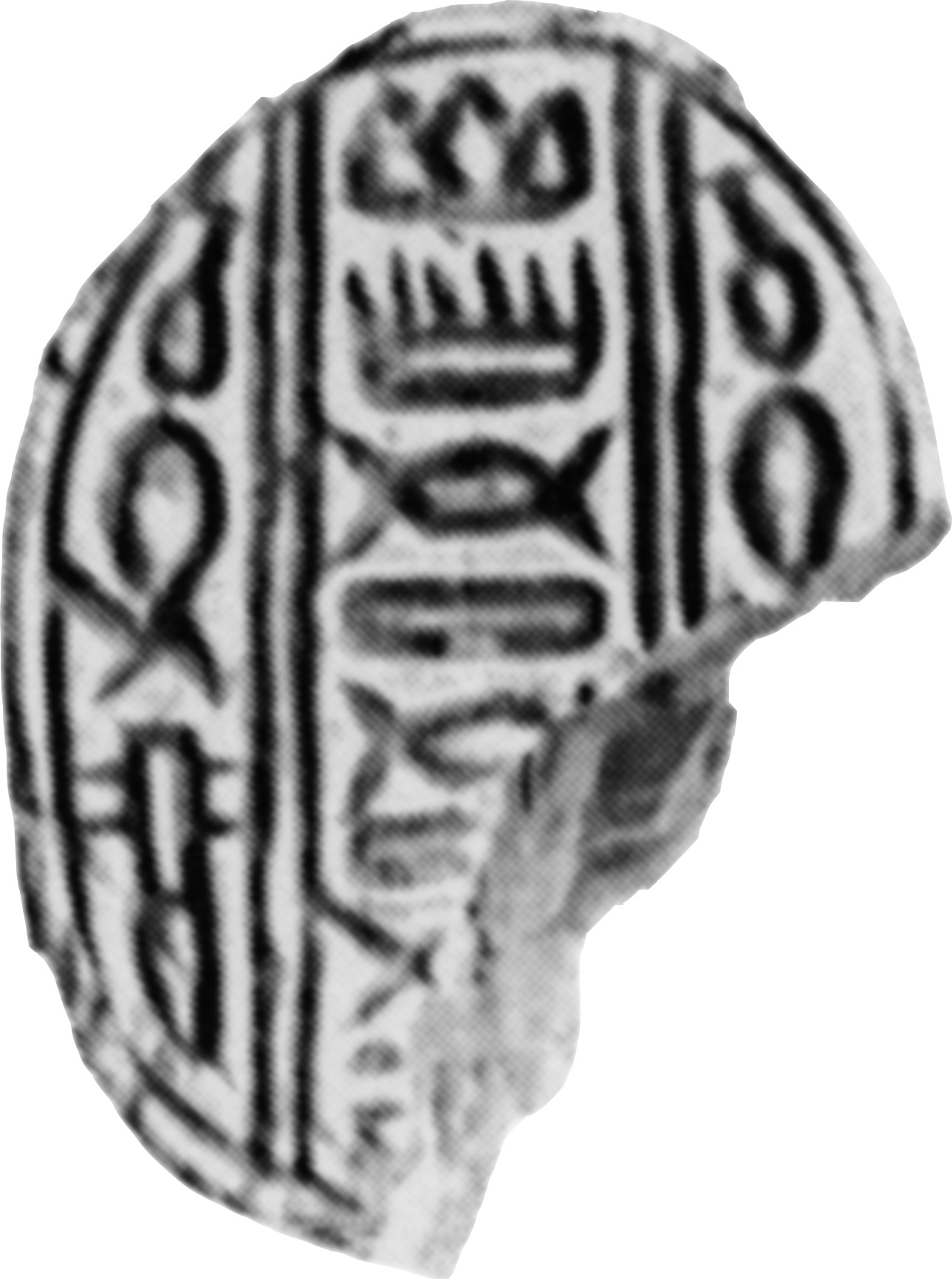} & \includegraphics[width=1.2cm, height=1.4cm]{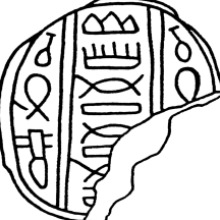} & 
  \includegraphics[width=1.2cm, height=1.4cm]{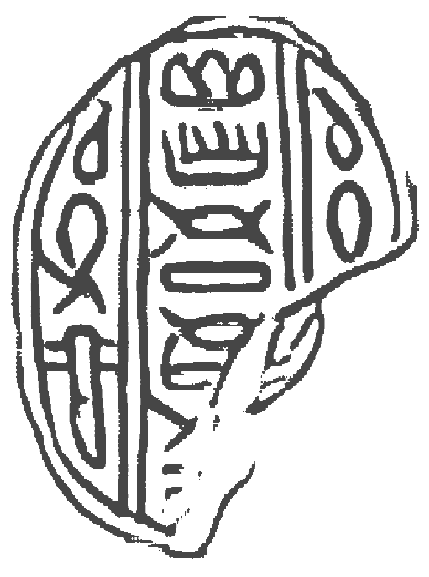} &
 \includegraphics[height=1.4cm, width=1.2cm]{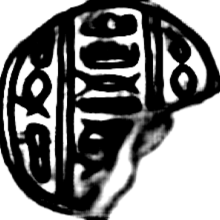} \\ 
 \multicolumn{4}{c}{sa}\\[0.1in]
(a) Input & (b) Manual & (c) Ours & (d) \cite{xsoria2020dexined}    
\end{tabular}
\caption{{\bf Image-to-drawing generation.} 
Our results are more similar to the manual drawings than those of~\cite{xsoria2020dexined} in generating delicate textures and in completing faint edges. 
Additional results are given in the supplemental materials.
}
\label{fig:generation}
\end{center}
\end{figure}

Figure~\ref{fig:generation} demonstrates our results qualitatively.
Since no prior work that generates drawings in our domain exists, to allow comparison we trained a SoTA supervised edge detector, {\em DexiNed}~\cite{xsoria2020dexined} on CSSL, considering paired drawings as edge maps.
A main challenge is to deal with noise, existing in images but not in drawings, caused as a result of the quality of the images, erosion of the artifacts, and the ability of the artist to complete the missing lines.
In addition, each artifact has defects, some of which also present in our generations. 
 {\em DexiNed}'s results miss the delicate textures and drawing features, such as the texture on the ibex's torso. 
DexiNed also generates thick edges, in accordance with the image itself.
Conversely, our method generates the delicate textures, nicely completes faint edges, and uses edge thickness as learned from the manual drawings.
A couple of archaeologists we consulted with consider the generated drawings as useful.

Table ~\ref{table: generation metrics} quantitatively compares our results to DexiNed~\cite{xsoria2020dexined}'s, using common edge detecting metrics: {\em F-measure} of {\em Optimal Dataset Scale (ODS)}, {\em Optimal Image Scale (OIS)} and {\em Average Precision (AP)}.
In this experiment we consider the manual drawings as ground truth edge maps.
Our results outperforms DexiNed's in all metrics.

We note that our model is even able to generate accurate drawings from photos of objects it was not trained on, such as artistic reliefs, sculptures and a variety of archaeological artifacts.
We show qualitative results in the supplementary.

\begin{table}
\centering
\begin{tabular}{ |l|c|c|c|  }
 \hline
 Training method & ODS & OIS & AP \\
 \hline
  \hline
 Our drawings    & {\bf0.38} & {\bf0.39} & {\bf0.29} \\
  \hline
 DexiNed~\cite{xsoria2020dexined}'s drawings   &0.28 & 0.28 & 0.25 \\
  \hline
\end{tabular}
\caption{{\bf Image-to-drawing generation quantitative results.} 
Our model outperforms DexiNed~\cite{xsoria2020dexined}'s in  
 common metrics for edge detection, considering the manual drawings as ground truth.}
\label{table: generation metrics}
\end{table}

\section{Ablation study}
\label{sec:ablation}
This section evaluates the benefits of the different components of our method.
In particular, it evaluates the contribution of the drawings for the training, of jointly solving classification and drawing generation, of training with both labeled and unlabeled data and of training two separate encoders.
We also examine the benefit of our method when the amount of labeled data is low, which is likely to be the case for future archaeological datasets. 
For that sake, we use three different sizes of labeled pairs during 2-fold cross validation training: full dataset ($507$,$513$), half dataset ($256$,$259$) and a quarter of the dataset ($129$,$132$).
 The size and split of the test set remains the same, $507$ \& $513$ pairs.
In all experiments we use the Resnet101 backbone. Similar results are obtained for other backbones.

\vspace{0.01in}
\noindent
{\bf The contribution of the drawings for the training.}
We trained our model with and without drawings.
Recall that the input at inference is always an image, as this is the prevalent available data.
Table~\ref{table: training w/wt drawings} shows that as expected,
training with pairs is indeed preferable.
Moreover the more data, the better.
Finally, it shows that the less available data during training, the more important it is to use paired data.
This is tested in two cases, when the omitted examples are not used at all and when the omitted examples are considered as unlabeled pairs.

\begin{table}
\centering
\begin{tabular}{ |l|c|c|c|  }
 \hline
 Input type / size & Full set  & $1/2$ set & $1/4$ set \\
 \hline
  \hline
 Ours: omitted unlabeled    &89.6\% & 86.5\% &78.0\%\\
  \hline
Ours: omitted unused    &89.6\% & 85.8\% &77.8\%\\
  \hline
 Photos only    &82.5\% &72.0\% &60.3\%\\
  \hline
\end{tabular}
\caption{
{\bf Training with different inputs.}
The classification results, when training with paired images \& drawings, outperform the results when using only images during training. 
Furthermore, the less data available, the more important it is to use these pairs.
}
\label{table: training w/wt drawings}
\end{table}

\vspace{0.01in}
\noindent
{\bf The benefit of jointly solving all tasks.}
We evaluate the impact on classification of image-drawing similarity and of drawing generation.
Toward this end, we checked the impact of \(\mathcal{L}_{Sim}\) and \(\mathcal{L}_{Gen}\).
Specifically, we trained the classification model without  \(\mathcal{L}_{Gen}\) and then a model that solves generation without forcing embedding similarity \(\mathcal{L}_{Sim}\) (Equation~\ref{eq:stage_1_loss}). 
Table~\ref{table: training w/wt generation and similarity} shows that, as expected,      \(\mathcal{L}_{Sim}\) is crucial for classification. 
Furthermore, adding \(\mathcal{L}_{Gen}\)  improves classification as well, while providing drawings that are important for documentation.

\begin{table}
\centering
\begin{tabular}{ |l|c|c|c|  }
 \hline
 Training set size & Full set & $1/2$ set & $1/4$ set \\
 \hline
  \hline
 Full method    &89.6\% & 86.5\% &78.0\%\\
  \hline
   w/o \(\mathcal{L}_{Sim}\)    & 82.2\% & 72.1\% & 61.8\%\\
  \hline 
 w/o \(\mathcal{L}_{Gen}\)    & 87.8\% & 84.1\% & 73.9\%\\
  \hline
\end{tabular}
\caption{{\bf Impact of \(\mathcal{L}_{Sim}\) and \(\mathcal{L}_{Gen}\) on classification.}
Both \(\mathcal{L}_{Sim}\)  \(\mathcal{L}_{Gen}\)
 are crucial for the accuracy of solving classification.
 }
\label{table: training w/wt generation and similarity}
\end{table}

\vspace{0.01in}
\noindent
{\bf Semi-supervised vs. fully supervised.}
This experiment studies the impact of using unlabeled data.
We trained two models, 
one only with the available labeled data (fully supervised) and the other also with the unlabeled data.
In the first case we used $1,020$ labeled pairs, and in the second we used the additional $5,616$ unlabeled pairs.

Table~\ref{table: Semi-supervised vs. Supervised} shows that for all sizes of training sets, the results achieved by the semi-supervised training outperforms those of  the supervised training.
Thus, additional input, even if unlabeled, should be used.
Furthermore, the fewer the labeled data is, the more beneficial semi-supervision  is. 
This is so since when having fewer labeled pairs, but the same number of unlabeled pairs, their influence grows.

\vspace{0.01in}
\noindent
{\bf Shared encoder vs. separate encoders.}
Our approach employs 2 encoders and achieves accuracy of $89.6\%$.
If instead we used a shared encoder the accuracy decreases to $85.15\%$.

\vspace{0.01in}
\noindent
{\bf Hyper-parameters.}
These are quite robust.
Specifically, in the loss function,
if we change the most important $\gamma_1$ (similarity), which is $0.8$, by $\pm 0.1$ (at the expense of $\gamma_2$), the accuracy change will be limited to $0.5\%$.
Increasing \(\alpha\) over \(\beta\) affects the generation; however changing their values from $(0.3,0.7)$ to $(0.4,0.6)$ for instance, the impact will be almost invisible.
More details are given in the supplementary.

\begin{table}
\centering
\begin{tabular}{ |l|c|c|c|  }
 \hline
 Training  & Full set & $1/2$ set & $1/4$ set \\
 \hline
  \hline
 Ours: labeled+unlabeled    &89.6\% & 86.5\% &78.0\%\\
  \hline
 Ours: only labeled    &83.8\% & 75.8\% & 64.8\%\\
  \hline
\end{tabular}
\caption{{\bf Supervised vs. semi-supervised training.}
This experiment shows that unlabeled data is beneficial and improves the results.
Since we do not expect to have much labeled data in this domain, this is very important.
}
\label{table: Semi-supervised vs. Supervised}
\end{table}

\begin{figure}[t]
\begin{center}
\begin{tabular}{c c c}
    \includegraphics[width=2.3cm]{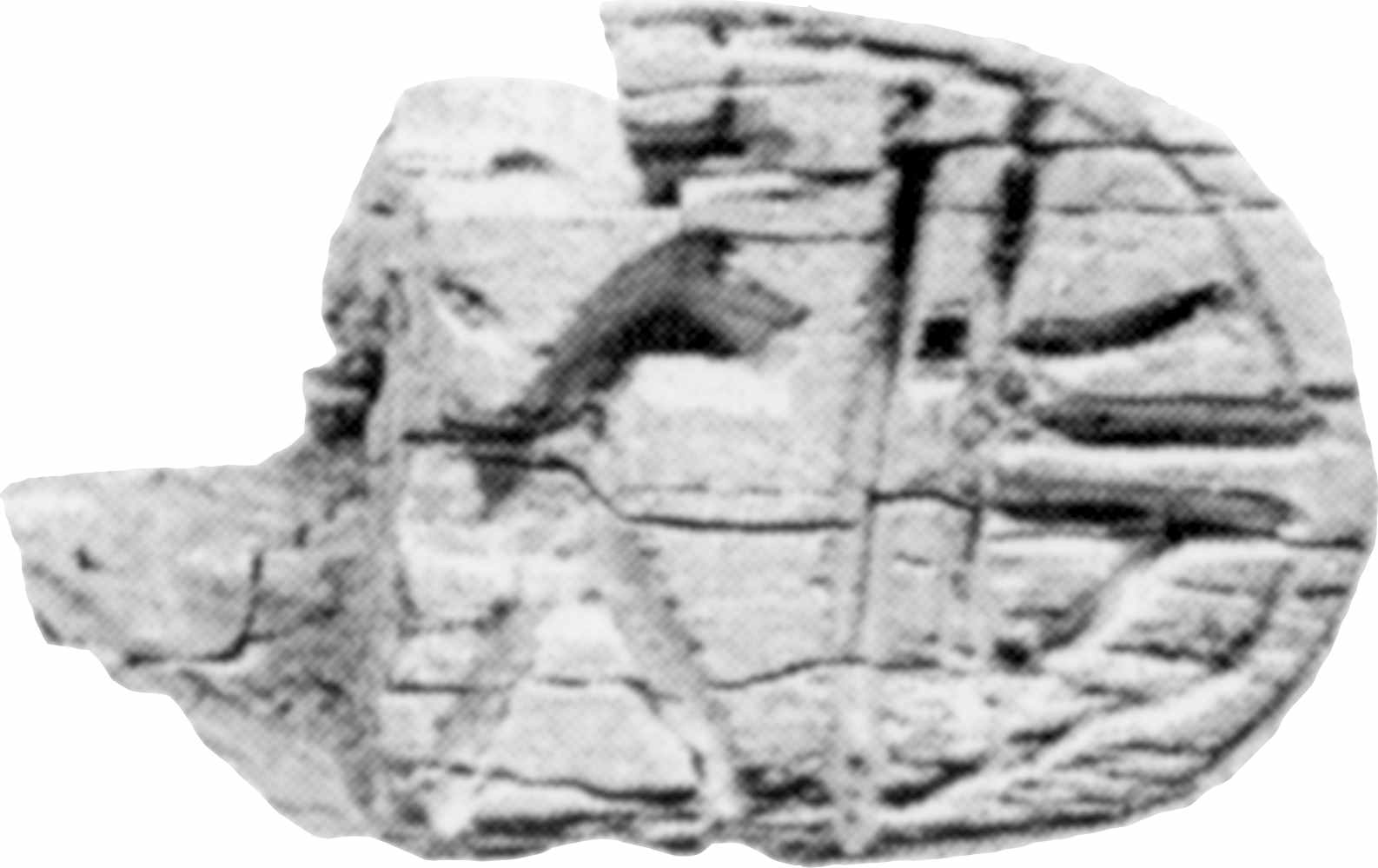} & \includegraphics[width=1.9cm]{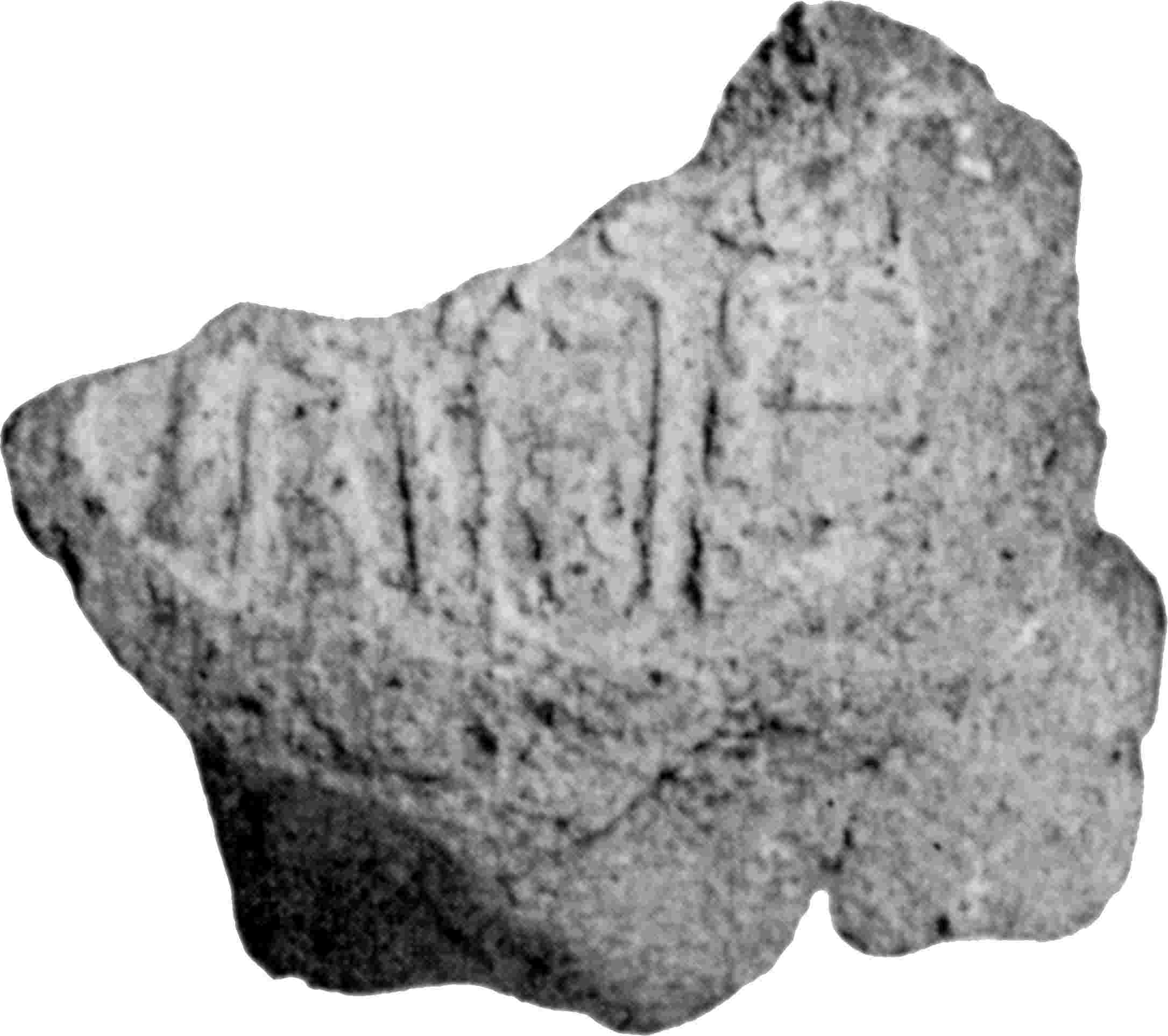} & 
    \includegraphics[width=1.8cm]{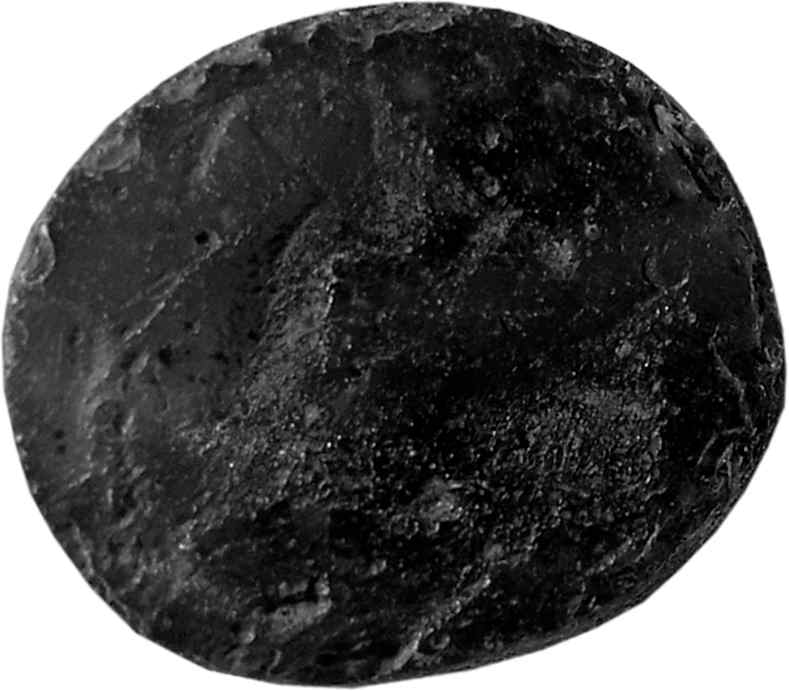}
     \\
     (a) lion  & (b) beetle & (c) ibex
\end{tabular}
\caption{{\bf Classification limitation.}
These three worn-out artifacts are classified erroneously: 
the lion as as anthropomorphic~(a), the beetle as ankh~(b) and the ibex as lion~(c).
}
\label{fig:class. limitation}
\end{center}
\begin{center}
\begin{tabular}{c c c c}
  \includegraphics[height=0.11\textwidth]{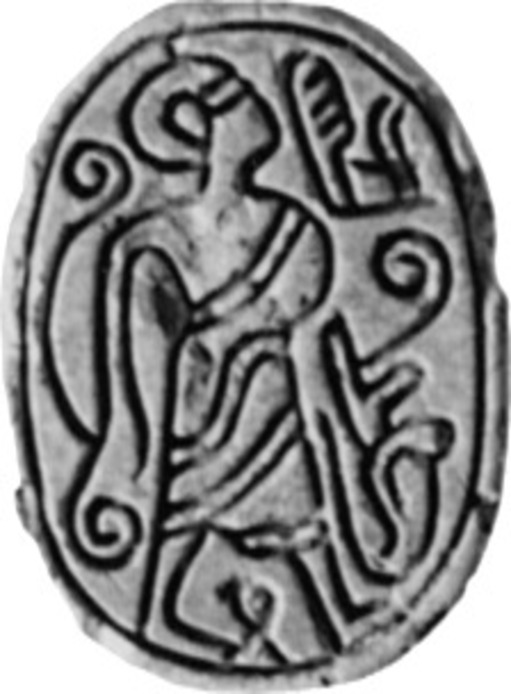} & \includegraphics[height=0.11\textwidth]{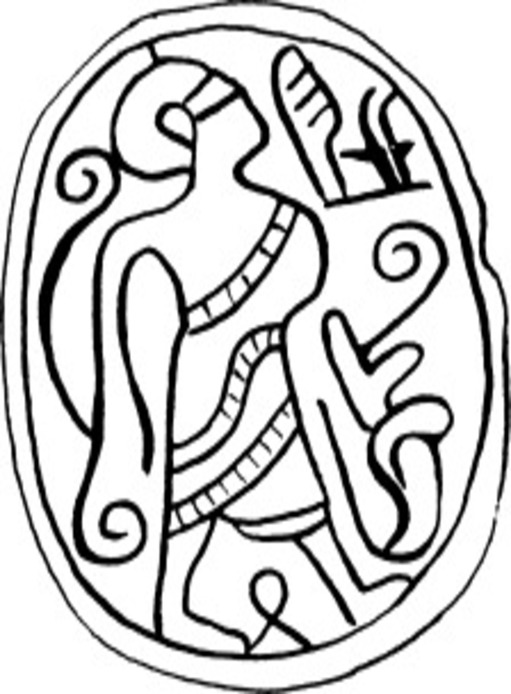} & 
  \includegraphics[height=0.11\textwidth]{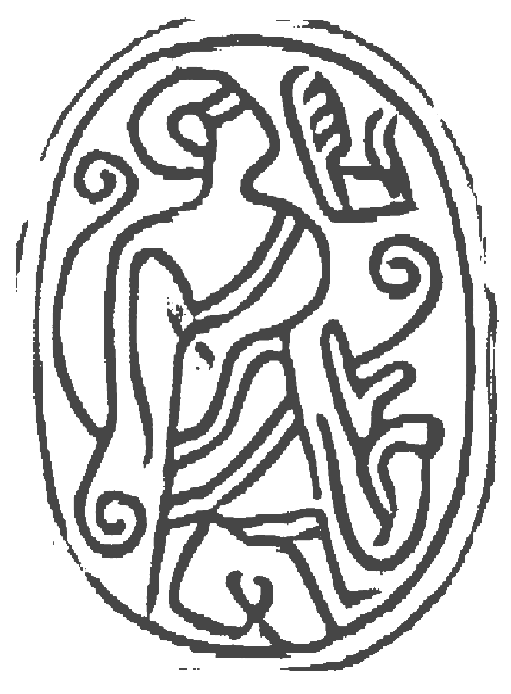} &
 \includegraphics[height=0.11\textwidth]{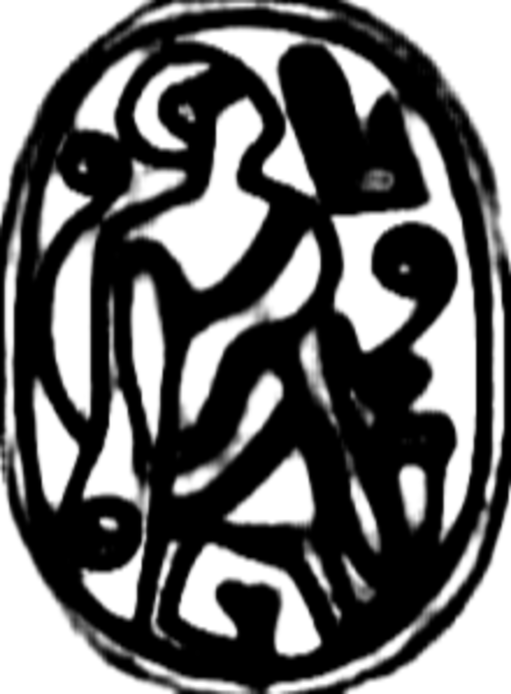} \\
 \includegraphics[height=0.07\textwidth]{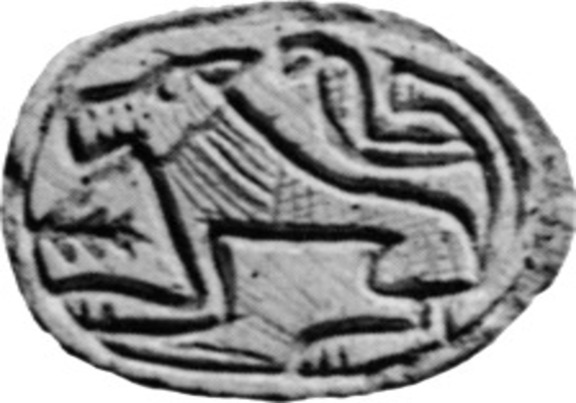} & \includegraphics[height=0.07\textwidth]{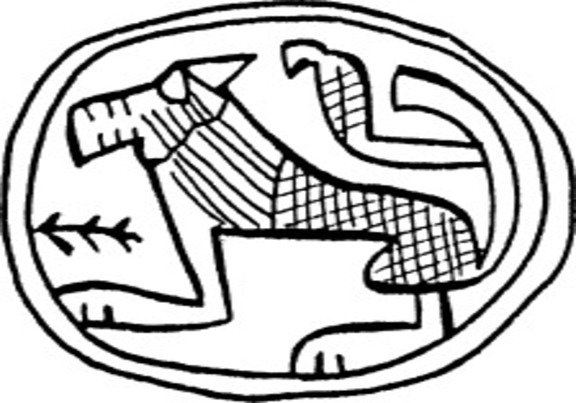} & 
  \includegraphics[height=0.07\textwidth]{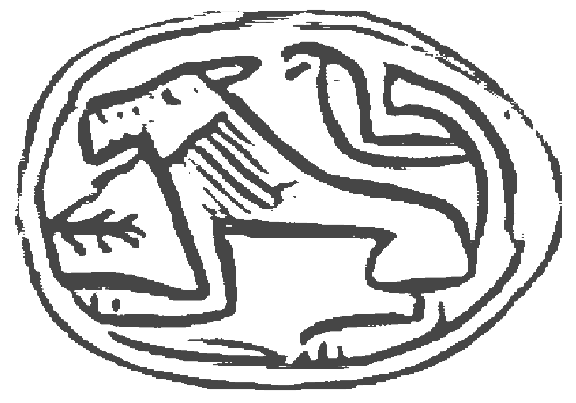} &
 \includegraphics[height=0.07\textwidth]{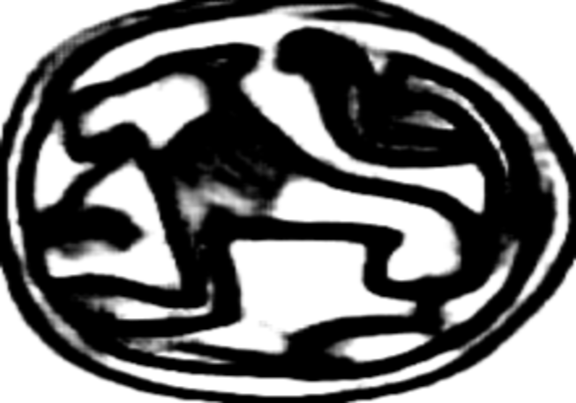} \\
 Input & Manual & Ours & \cite{xsoria2020dexined}  
\end{tabular}
\caption{{\bf Image-to-drawing limitation.} 
Our method might fail to draw some fine details, such as the buckles on the anthropomorphic belt and some decorations on the lion's body and tail.
Still, our results preserve most of the original details, compared to~\cite{xsoria2020dexined}. }
\label{fig:limitation}
\end{center}
\end{figure}

\vspace{0.01in}
\noindent
{\bf Limitations.}
Figure~\ref{fig:class. limitation} shows cases where our model fails to classify the objects correctly.
In these cases, the artifacts are worn out and are erroneously classified into related classes.

Figure~\ref{fig:limitation} shows cases where our generated drawings do not succeed to generate the delicate textures. 
It can be seen though, that our drawings are still better than those of ~\cite{xsoria2020dexined}.

\section{Conclusions}


In archaeology, artifacts are studied using their photographs. 
For a subset of them, drawings are made by trained drafts people.
These are experts in their field and are able to complete features that are not visible due to the artifacts eroded condition. 
The challenge addressed in this paper is how such data can be used to solve classification in the case of a small damaged dataset. 
We show that implicit knowledge obtained from a drawing is transferred to an image, guiding it to the more important features. 
Furthermore, we show that performing the training in a semi-supervised way, takes advantage of unlabelled image-drawing pairs. 
 
In addition, our model generates from the image a drawing of the object. 
This is challenging since the image and drawing are not exactly aligned and what is transferred is the approximate position of the image features. 
The resulting model is able to mimic with high accuracy the knowledge and the drawing expertise of the artist, as well as the knowledge of the archaeologist. 
Our method can be generalized to objects of reliefs, either archaeological or artistic.

Last but not least, we created a relatively large and challenging dataset, which can be used in future research.

\noindent
{\bf ACKNOWLEDGMENTS}. We gratefully acknowledge the support of the Israel Science Foundation (ISF) 1083/18 and Ministry Science and Technology (MOST)  3-17513.

{\small
\bibliographystyle{ieee_fullname}
\bibliography{egbib}
}

\end{document}